\def\year{2020}\relax
\documentclass[letterpaper]{article} % DO NOT CHANGE THIS
\usepackage{aaai20}  % DO NOT CHANGE THIS
\usepackage{times}  % DO NOT CHANGE THIS
\usepackage{helvet} % DO NOT CHANGE THIS
\usepackage{courier}  % DO NOT CHANGE THIS
\usepackage[hyphens]{url}  % DO NOT CHANGE THIS
\usepackage{graphicx} % DO NOT CHANGE THIS
\usepackage{booktabs} % for professional tables
\usepackage{microtype}
\usepackage{subfigure}

\usepackage{amsmath}
\usepackage{amsthm}
\usepackage{amssymb}
\usepackage{epstopdf}
\usepackage{multirow}
\usepackage{array}
\RequirePackage{algorithm}
\RequirePackage{algorithmic}
\newcommand{\citet}[1]{\citeauthor{#1} \shortcite{#1}}
\newcommand{\citep}{\cite}

\usepackage{graphicx}  % DO NOT CHANGE THIS
\title{Object-Oriented Dynamics Learning through Multi-Level Abstraction}
\author{Guangxiang Zhu\textsuperscript{\rm 1}\thanks{Equal contribution}, Jianhao Wang\textsuperscript{\rm 1}\footnotemark[1], Zhizhou Ren\textsuperscript{\rm 1}\footnotemark[1], Zichuan Lin\textsuperscript{\rm 2}, Chongjie Zhang\textsuperscript{\rm 1}\\
	\textsuperscript{\rm 1}Institute for Interdisciplinary Information Sciences, Tsinghua University, Beijing, China\\
	\textsuperscript{\rm 2}Department of Computer Science and Technology, Tsinghua University, Beijing, China\\
\{zhu-gx15, wjh19, rzz16, linzc16\}@mails.tsinghua.edu.cn, chongjie@tsinghua.edu.cn % email address must be in roman text type, not monospace or sans serif
}
 
\begin{document}

\maketitle

\begin{abstract}
Object-based approaches for learning action-conditioned dynamics has demonstrated promise for generalization and interpretability. However, existing approaches suffer from structural limitations and optimization difficulties for common environments with multiple dynamic objects. In this paper, we present a novel self-supervised learning framework, called \emph{Multi-level Abstraction Object-oriented Predictor} (MAOP), which employs a three-level learning architecture that enables efficient object-based dynamics learning from raw visual observations. We also design a spatial-temporal relational reasoning mechanism for MAOP to support instance-level dynamics learning and handle partial observability. Our results show that MAOP significantly outperforms previous methods in terms of sample efficiency and generalization over novel environments for learning environment models. We also demonstrate that learned dynamics models enable efficient planning in unseen environments, comparable to true environment models. In addition, MAOP learns semantically and visually interpretable disentangled representations.
\end{abstract}

\section{Introduction}
Model-based deep reinforcement learning (DRL) has recently attracted much attention for improving sample efficiency of DRL \citep{gu2016continuous,racaniere2017imagination,finn2017deep}. One of the core problems for model-based DRL is to learn action-conditioned dynamics models through interacting with environments. Pixel-based approaches have been proposed for such dynamics learning from raw visual perception, achieving remarkable performance in training environments \citep{oh2015action,chiappa2017recurrent}.

To unlock sample efficiency of model-based DRL, learning action-conditioned dynamics models that generalize over unseen environments is critical yet challenging. \citet{finn2016unsupervised} proposed a dynamics learning method that takes a step towards generalization over object appearances. \citet{zhu2018object} developed an object-oriented dynamics predictor to support generalization. However, due to structural limitations and optimization difficulties, these methods do not efficiently generalize over environments with multiple controllable and uncontrollable dynamic objects and different static object layouts.

To address these limitations, we propose a novel three-level learning framework for self-supervised learning of object-oriented dynamics model, called \emph{Multi-level Abstraction Object-oriented Predictor} (MAOP). This framework simultaneously learns disentangled object representations and predicts object motions conditioned on their historical states, their interactions to other objects, and an agent's actions. To reduce the complexity of such concurrent learning and improve sample efficiency, MAOP employs a three-level learning architecture from the most abstract level of motion detection, to dynamic instance segmentation, and to dynamics learning and prediction. A more abstract learning level solves an easier problem and has lower learning complexity, and its output provides a coarse-grained guidance for a less abstract learning level, improving its speed and quality of learning. 

Specifically, we perform motion detection to detect proposal regions that potentially contain dynamic instances for the follow-up dynamic instance segmentation. Then we exploit spatial-temporal information of locomotion property and appearance patterns to capture coarse region proposals of dynamic instances. Finally we use them to guide the learning of the object representations and instance localization at the level of dynamics learning. This three-level architecture is inspired by humans' multi-level motion perception from cognitive science studies \citep{johansson1975visual,lu1995functional} and multi-level abstraction search in constraint optimization \citep{zhang2016co}. In addition, we design a novel CNN-based spatial-temporal relational reasoning mechanism for MAOP, which includes a Relation Net to reason about spatial relations between objects and an Inertia Net to learn temporal effects. This mechanism offers a disentangled way to handle physical reasoning in settings with partial observability.

Our results show that MAOP significantly outperforms previous methods for learning dynamics models in terms of sample efficiency and generalization over novel settings with multiple controllable and uncontrollable dynamic objects and different object layouts. MAOP enables model learning from few interactions with environments and accurately predicting the dynamics of objects as well as raw visual observations in previously unseen environments. The learned dynamics model enables an agent to directly plan in unseen environments without retraining. In addition, MAOP learns disentangled representations and gains visually and semantically interpretable knowledge, including meaningful object masks, accurate object motions, disentangled relational reasoning process, and controllable factors. Last but not least, MAOP provides a general multi-level framework for learning object-based dynamics model from raw visual observations, offering opportunities to easily leverage well-studied object detection methods (e.g., Mask R-CNN \citep{he2017mask}) in the area of computer vision.

\section{Related Work}
\textbf{Object-oriented reinforcement learning} has received much research attention, which exploits efficient representations based on objects and their interactions. This learning paradigm is close to that of human cognition in the physical world and the learned object-level knowledge can be efficiently generalized across environments. Early work on object-oriented RL requires explicit encodings of object representations, such as \emph{relational MDPs} \citep{guestrin2003generalizing}, \emph{OO-MDPs} \citep{diuk2008object}, object focused q-learning \citep{cobo2013object}, and Schema Networks \citep{kansky2017schema}. In this paper, we present an end-to-end, self-supervised neural network framework that automatically learns object representations and dynamics conditioned on actions and object relations from raw visual observations.

\textbf{Action-conditioned dynamics learning} aims to address one of the core problems for model-based DRL, i.e., constructing an environment dynamics model. Several pixel-based approaches have been proposed for learning how an environment changes in response to actions through unsupervised video prediction and achieve remarkable performance in training environments \citep{oh2015action,chiappa2017recurrent}. \citet{fragkiadaki2015learning} propose an object-centric prediction method to learn the dynamics model when given the object localization and tracking. \citet{finn2016unsupervised} propose an action-conditioned video prediction method that explicitly models pixel motion and learns invariance to object appearances. Recently, \citet{zhu2018object} propose an object-oriented dynamics learning paradigm. However, it focuses on environments with a single dynamic object. In this paper, we take a further step towards object-oriented dynamics modeling in more general environments with multiple dynamic objects and also demonstrate its usage for model-based planning. In addition, we design an instance-aware dynamics mechanism to support instance-level dynamics learning and handle partial observations.

\textbf{Relation-based deep learning approaches} make significant progress in a wide range of domains such as physical reasoning \citep{chang2016compositional,battaglia2016interaction,van2018relational}, computer vision \citep{watters2017visual,wu2017learning}, natural language processing \citep{santoro2017simple}, and reinforcement learning \citep{zambaldi2018relational,zhu2018object}. Relation-based nets introduce relational inductive biases into neural networks, which facilitate generalization over entities and relations and enables relational reasoning \citep{battaglia2018relational}. This paper proposes a novel spatial-temporal relational reasoning mechanism, which includes a CNN-based Inertia Net for learning temporal effects in addition to a CNN-based Relation Net for reasoning about spatial relations.

\textbf{Instance Segmentation} has been one of the fundamental problems in computer vision and many approaches have been proposed \citep{pinheiro2015learning,li2016fully,he2017mask}. Instance segmentation can be regarded as the combination of semantic segmentation and object localization.  Most models are supervised learning and require a large labeled training dataset. \citet{liu2015multi} proposes a weakly-supervised approach to infer object instances in foreground by exploiting dynamic consistency in video. In this paper, we design a self-supervised, three-level approach for learning dynamic rigid object instances. First, foreground detection computes region proposals for potential dynamic objects. Based on these region proposals, we then learn coarse dynamic instance segmentation. This coarse instance segmentation provides a guidance for learning accurate instances at the dynamics learning level, whose instance segmentation considers not only object appearances but also motion prediction conditioned on object-to-object relations and actions.

\begin{figure*}[ht]
	\begin{center}

		\centering
		\includegraphics[width=1.8\columnwidth]{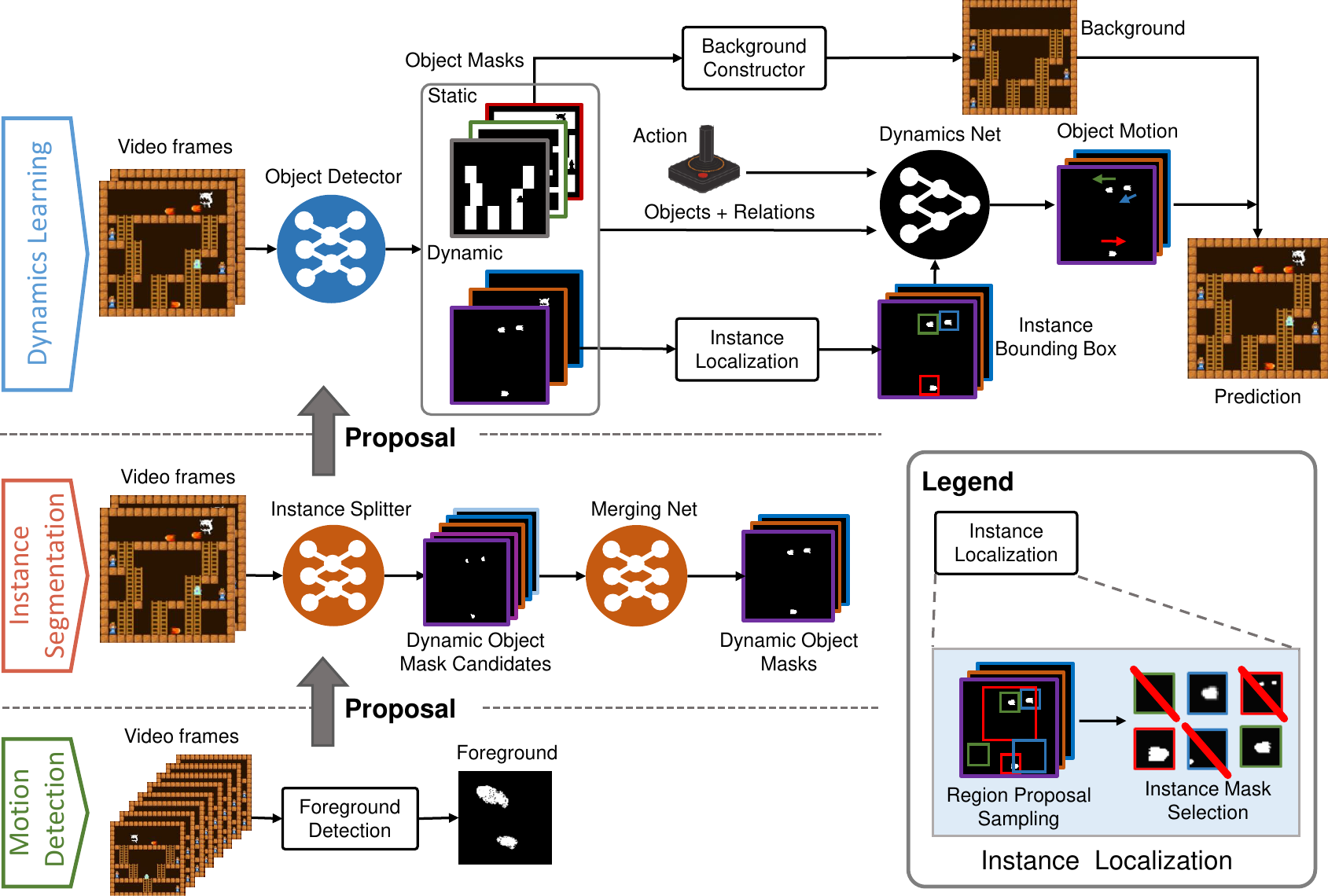}

		\caption{Multi-level dynamics learning framework. From a bottom-up view, we first perform motion detection to produce foreground masks. Then, we utilize the foreground masks as dynamic region proposals to guide the learning of dynamic instance segmentation. Finally, we use the learned dynamic instance segmentation networks (Instance Splitter and Merging Net) as a guiding network to generate region proposals of dynamic instances and guide the learning of Object Detector at the level of dynamics learning. We provide a pseudocode that sketches out this multi-level framework in Algorithm \ref{pseudocode}. }
		\label{overview}
	\end{center}

\end{figure*}

\section{Multi-level Abstraction Object-Oriented Predictor (MAOP)}
In this section, we will present a novel self-supervised deep learning framework, aiming to learn object-oriented dynamics models that are able to efficiently generalize over unseen environments with different object layouts and multiple dynamic objects. Such a generalized object-oriented dynamics learning approach requires simultaneously learning object representations and motions conditioned on their historical states, their interactions to other objects, and an agent's actions. This concurrent learning is challenging for an end-to-end approach in complex environments. Evidences from cognitive science studies \citep{johansson1975visual,lu1995functional} show that human beings are born with prior motion perception ability (in Cortical area MT) of perceiving moving and motionlessness, which enables learning more complex knowledge, such as object-level dynamics prediction. Inspired by these studies, we design a multi-level learning framework, called \emph{Multi-level Abstraction Object-oriented Predictor} (MAOP), which incorporates motion perception levels to assist in dynamics learning. Since we focus on the seminal study towards the multi-level framework for interpretable and efficient dynamics learning, we make a basic assumption that the environment only contains rigid objects, which has also been widely adopted by many papers \citep{watters2017visual,wu2017learning,zhu2018object}. 

Figure \ref{overview} illustrates three levels of MAOP framework: \emph{dynamics learning}, \emph{dynamic instance segmentation}, and \emph{motion detection}. Here we present them from a top-down decomposition view. The \emph{dynamics learning} level is an end-to-end, self-supervised neural network, aiming to learn object representations and instance-level dynamics, and predict the next visual observation conditioned on object-to-object relations and an agent's action. To guide the learning of the object representations and instance localization at the level of dynamics learning, the more abstracted level of \emph{dynamic instance segmentation} learns a guiding network in a self-supervised manner, which can provide coarse mask proposals of dynamic instances. This level exploits spatial-temporal information of locomotion property and appearance patterns to capture region proposals of dynamic instances. To facilitate the learning of instance segmentation, MAOP employs the more coarse-grained level of \emph{motion detection}, which detects changes in image sequences and provides guidance on proposing regions potentially containing dynamic instance. 

Algorithm \ref{pseudocode} shows a pseudocode that summarizes the training process of our framework. As the learning proceeds, the knowledge distilled from the more coarse-grained level are gradually refined at the more fine-grained level by considering additional information. When the training is finished, the coarse-grained levels of dynamic instance segmentation and motion detection will be removed at the testing stage. In the rest of this section, we will describe in detail the design of each level and their connections.

\begin{algorithm*}[htbp]
	\caption{Training process for our multi-level framework.}
	\begin{algorithmic}[1]
		\STATE Initialization. Initialize the parameters of all neural networks with random weights respectively.
		\STATE Motion Detection Level. Perform foreground detection to produce dynamic region proposals, which potentially have moving objects.
		\STATE Instance Segmentation Level. Train the dynamic instance segmentation network (including Instance Splitter and Merging Net) by minimizing $\mathcal{L}_{\text{DIS}}$, which includes a proposal loss to focus the dynamic instance segmentation on the dynamic region proposals from Step 2.
		\STATE Dynamic learning Level. Train the dynamics learning network by minimizing $\mathcal{L}_{\text{DL}}$, which includes a proposal loss to utilize the dynamic instance proposals generated by the trained dynamic instance segmentation network in Step 3 to facilitate the learning of Object Detector.
	\end{algorithmic}
	\label{pseudocode}
\end{algorithm*}

\subsection{Motion Detection Level}
At this level, we employ foreground detection to detect potential regions of dynamic objects from a sequence of image frames and provide coarse dynamic region proposals for assisting in dynamic instance segmentation. In our experiments, we use a basic unsupervised foreground detection approach \citep{lo2001automatic}. Our framework is also compatible with many advanced unsupervised foreground detection methods \citep{lee2005effective,maddalena2008self,zhou2013moving,guo2014robust} that are more efficient or more robust to moving camera. These complex foreground detection methods have the potential to improve the performance but are not the focus of this work.

\subsection{Dynamic Instance Segmentation Level}\label{DIPN}
This level aims to generate region proposals of dynamic instances to guide the learning of object masks and facilitate instance localization at the level of dynamics learning. The architecture is shown in the middle level of Figure \ref{overview}. Instance Splitter aims to identify regions, each of which potentially contains one dynamic instance. As we focus on the motion of rigid objects, the affine transformation is approximatively consistent for all pixels of each dynamic instance mask. Inspired by this, we define a discrepancy loss $\mathcal{L}_{\text{instance}}$ for a sampled region that measures motion consistence of its pixels and use it to train Instance Splitter. To compute this loss, we first compute an average rigid transformation of a sampled region on object masks between two time steps, then apply this transformation to this region at the previous time step by Spatial Transformer Network (STN) \citep{jaderberg2015spatial}, and finally compared this predicted region with the region at the current time (the difference is measured by $l_2$ distance). Obviously, when a sampled region contains exactly one dynamic instance, this loss will be extremely small, and even zero when object masks are perfectly learned. As $\mathcal{L}_{\text{instance}}$ decreases on every sampled regions of object masks, Instance Splitter gradually learns to isolate dynamic instances from background and divide different dynamic objects onto different masks.

Considering that one object instance may be split into smaller patches on different masks, we append a Merging Net (i.e., a two-layer CNN with 1 kernel size and 1 stride) to Instance Splitter to learn to merge masks. This module uses a merging loss  $\mathcal{L}_{\text{merge}}$ that aims to merge mask candidates that are adjacent and share the same motion. In addition, we add a foreground proposal loss $\mathcal{L}_{\text{forground}}$ to encourage attentions on dynamic regions provided by the level of motion detection, which is defined similar to $\mathcal{L}_{\text{proposal}}$ at the level of dynamics learning. The total loss of this level is given by,
$
\mathcal{L}_{\text{DIS}}=\mathcal{L}_{\text{instance}} + \lambda_{3}\mathcal{L}_{\text{merge}}+ \lambda_{4}\mathcal{L}_{\text{forground}}.
$

Although the network structure of this level is similar to Object Detector in the level of dynamics learning, we do not integrated them together as a whole network because concurrent learning of both object representations and dynamics model is not stable. Instead, we first learn the coarse object representations only based on the spatial-temporal consistency of locomotion and appearance pattern, and then use them as proposal regions to guide object-oriented dynamics learning at the more fine-grained level, which in turn fine-tunes the object representations. In addition, MAOP is also readily to incorporate Mask R-CNN \citep{he2017mask} or other off-the-shelf supervised or unsupervised object detection methods \citep{liu2018deep,xu2019unsupervised} as a plug-and-play module into our framework to generate region proposals of dynamic instances.

\subsection{Object-Oriented Dynamics Learning Level}\label{IL}
The semantics of this level is formulated as learning an object-based dynamics model with region proposals generated from the more abstracted level of dynamic instance segmentation. Its architecture is shown at the top level of Figure \ref{overview}, which is an end-to-end neural network and can be trained in a self-supervised manner. It takes a sequence of video frames and an agent's actions as input, learns disentangled representations (including objects, relations and effects) and dynamics of controllable and uncontrollable dynamic object instances conditioned on actions and object relations, and produces predictions of future frames. The whole architecture includes four major components: A) an Object Detector that learns to decompose the input image into objects; B) an Instance Localization module responsible for localizing dynamic instances; C) a Dynamics Net for learning the motion of each dynamic instance conditioned on the effects from actions and object-level spatial-temporal relations; and D) a Background Constructor that computes a background image from learned static object masks. In addition to Figure \ref{overview}, we further provide Algorithm \ref{aloodl} in Appendix to describe interactions of these components and the learning paradigm of object-based dynamics, which is a general framework and agnostic to the concrete form of each component. In the following paragraphs, we describe detailed design of each component.

\begin{figure*}[ht]
	\begin{center}
		
		\centering
		\includegraphics[width=1.7\columnwidth]{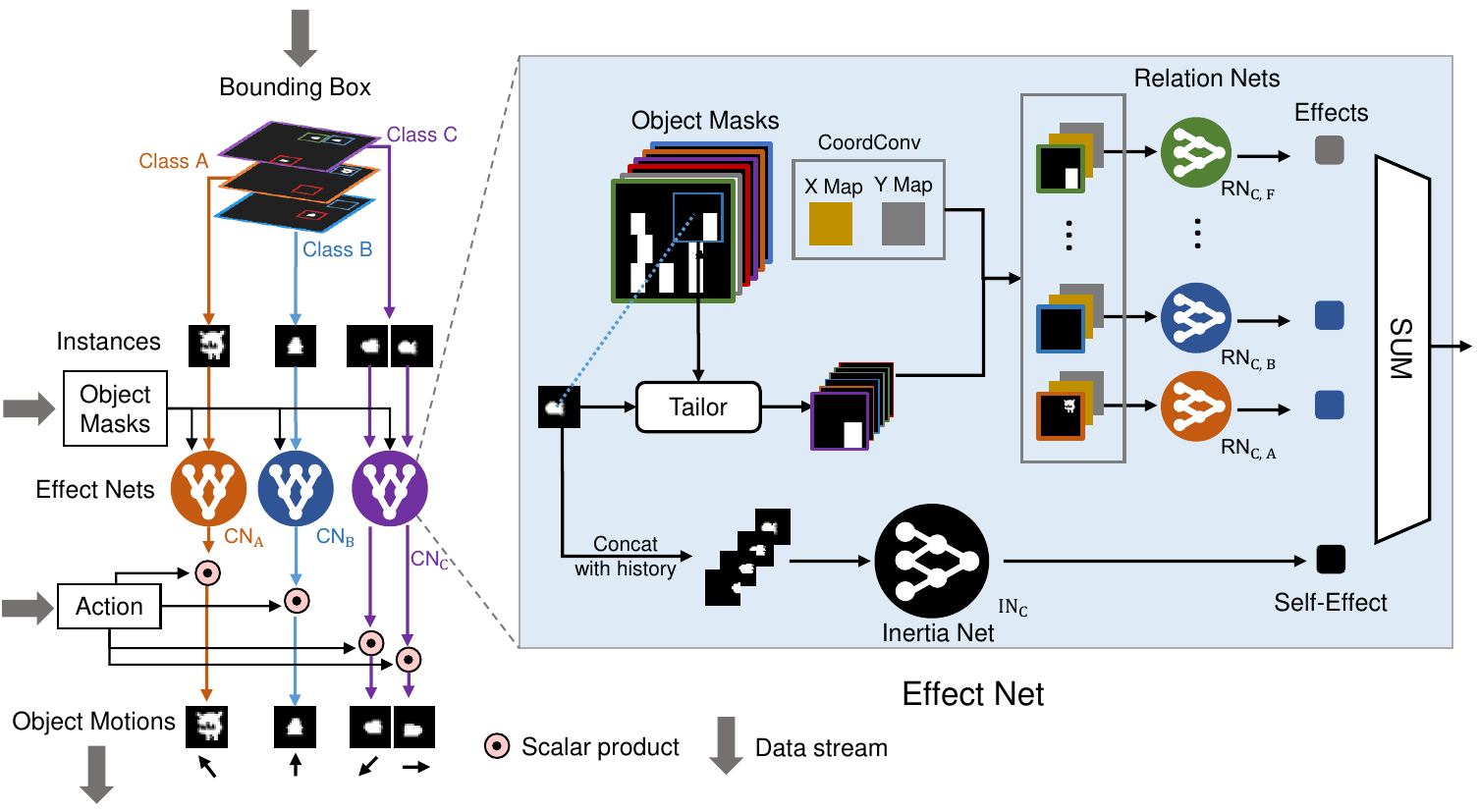}
		
		\caption{Architecture of Dynamics Net (left) and its component Effect Net (right). Different classes of objects are distinguished by different letters (e.g., A, B, ... , F). Dynamics Net has one Effect Net for each class of objects. An Effect Net consists of one Inertia Net and several Relation Nets.}
		\label{DynamicsNet}
	\end{center}
	
\end{figure*}

\textbf{Object Detector and Instance Localization Module.}\label{ODIL}
Object Detector is a CNN module aiming to learn object masks from a sequence of input images. An object mask describes a spatial distribution of a class of objects, which forms the fundamental building block of our object-oriented framework. Considering that instances of the same class are likely to have different motions, we append an Instance Localization Module to Object Detector to localize each dynamic instance to support instance-level dynamics learning. Class-specific object masks in conjunction with instance localization bridge visual perception (Object Detector) with dynamics learning (Dynamics Net), which allows learning objects based on both appearances and dynamics.

Specifically, Object Detector receives image $I^{t} \in \mathbb{R}^{H \times W \times 3}$ at timestep $t$ and outputs object masks $O^{t} \in [0, 1]^{H \times W \times n_O } $, including dynamic object masks $D^{t} \in [0, 1]^{ H \times W \times n_D}$ and static object masks $S^{t} \in [0, 1]^{ H \times W \times n_S} $, where $H$, $W$ denote the height and width of the input image, $n_D$ and $n_S$ denotes the maximum possible number of dynamic and static object classes respectively, and $n_O=n_D+n_S$. Note that our framework does not require the actual number of object classes, but needs to set a maximum number (usually 10 is enough). When they do not match, some learned object masks may be redundant, which does not affect the accuracy of predictions. We have conducted experiments to confirm this and will add the results. Entry $O_{u,v,i}$ indicates the probability that pixel $I_{u,v,:}$ belongs to the $i$-th object class. The Instance Localization module uses learned dynamic object masks to identify each object instance mask $X_{:,:,i}^{t} \in [0, 1]^{H_M \times W_M  } (1\le i \le n_M)$, where $H_M$, $W_M$ denote the height and width of the bounding box of this instance and $n_M$ denotes the maximum possible number of localized instances. As shown in Figure \ref{overview}, Instance Localization first samples a number of bounding boxes on dynamic object masks and then select the regions, each of which contains only one dynamic instance. We use the discrepancy loss $\mathcal{L}_{\text{instance}}$ described in Section of Dynamic Instance Segmentation Level as a selection score for selecting instance masks. More details of region proposal sampling and instance mask selection can be found in Appendix.

\textbf{Dynamics Net.}
Dynamics Net is designed to learn instance-level motion effects of actions, object-to-object spatial relations (Relation Net) and temporal relations of spatial states (Inertia Net), and to reason about the motion of each dynamic instance based on these effects. Its architecture is illustrated as Figure \ref{DynamicsNet}, where the motion of each dynamic instance is individually computed. We take as an example the computation of the motion of the $i$-th instance $X_{:,:,i}^{t}$ to illustrate the detailed structure of the Effect Net.

As shown in the right subfigure of Figure \ref{DynamicsNet}, Effect Net first uses a sub-differentiable tailor module introduced by \citet{zhu2018object} to enable the inference of dynamics focusing on the relations with neighbour objects. This module crops a $w$-size ``horizon'' window from the concatenated masks of all objects $O^{t}$ centered on the expected location of $X_{:,:,i}^{t}$, where $w$ denotes the maximum effective range of relations.  Then, the cropped object masks are concatenated with constant x-coordinate and y-coordinate meshgrid map (to make networks more sensitive to the spatial information) and fed into corresponding Relation Nets (RN) according to their classes. We use $C_{:,:,i,j}^{t}$ to denote the cropped mask that crops the $j$-th object class $O_{:,:,j}^{t}$ centered on the expected location of the $i$-th dynamic instance (the class it belongs to is denoted as $c_{i}, 1\le c_{i} \le n_D$). The effect of object class $j$ on class $c_{i}$, 
$E^{t}(c_{i},j)=\text{RN}_{c_{i},j}\Big(\text{concat}\big(C_{:,:,i,j}^{t},\text{Xmap},\text{Ymap}\big)\Big).
$
Note that there are total $n_D \times n_O $ RNs for $n_D \times n_O $ pairs of object classes that share the same architecture but not their weights. To handle the partial observation problem, we add an Inertia Nets (IN) to learn spatio-temporal self-effects of an object class through historical states,
$
E_{\text{self}}^{t}(c_{i})=\text{IN}_{c_{i}}\Big( \text{concat}\big( X_{:,:,i}^{t}, X_{:,:,i}^{t+1}, \dots , X_{:,:,i}^{t+h} \big) \Big),
$
where $h$ is the history length. There are total $n_D $ INs for $n_D $  dynamic object classes, which share the same architecture but not their weights. To predict the motion vector $m_{i}^{t}$ for the $i$-th dynamic instance, all these effects are summed up and then multiplied by the coded action vector $a^{t}$, that is,
$
m_{i}^{t}= \Big(\big(\sum_{j} E^{t}(c_{i},j)\big) + E_{\text{self}}^{t}(c_{i}) \Big) \cdot a^{t}.
$

\textbf{Background Constructor.}
This module constructs the static background of an input image based on the static object masks learned by Object Detector.
Since Object Detector can decompose its observation into objects in an unseen environment with different layouts, Background Constructor is able to generate a corresponding static background and support the visual observation prediction in novel environments. Specifically, Background Constructor maintains a background memory $B \in \mathbb{R}^{H \times W \times 3}$ which is continuously updated with the static object mask learned by Object Detector. Denoting $\alpha$ as the decay rate, the updating formula is given by, $B^{t}=\alpha B^{t-1} +(1-\alpha) I^{t} \sum_i S_{:,:,i}^{t}, \text{and} \ B^{0}=0$.

%\begin{equation*}
% B^{t}=\left\{
%\begin{aligned}
% &\alpha B^{t-1} +(1-\alpha) I^{t} \sum_i S_{:,:,i}^{t} \ ,  &      t>0 ;\\
%&0  ,   & t=0,
%\end{aligned}
%\right.
%\end{equation*}

\textbf{Prediction and Training Loss.}
Based on the learned masks and motions of the object instances, we propose an object-oriented prediction loss, $\mathcal{L}_{\text{pred-object}}= \sum_i \big\Arrowvert \text{STN} \Big( (\bar{u}_{i},\bar{v}_{i})^{t} , m_{i}^{t} \Big) - (\bar{u}_{i},\bar{v}_{i})^{t+1} \big\Arrowvert_2^2$, where $ (\bar{u}_{i},\bar{v}_{i})^{t}$ is the excepted location of $i$-th instance mask $X_{:,:,i}^{t}$. To utilize the information of ground-true future frames, we also use a conventional image prediction loss. Our prediction of the next frame is produced by merging the learned object motions and the background $B^{t}$. The pixels of a dynamic instance can be calculated by masking the raw image with the corresponding instance mask and we can use STN to apply the learned instance motion vector $m_{i}^{t}$ on these pixels. First, we transform all the dynamic instances according to the learned instance-level motions. Then, we merge all the transformed dynamic instances with the background image calculated from Background Constructor to generate the prediction of the next frame.
We use the pixel-wise $l_2$ loss to restrain image prediction error, denoted as $\mathcal{L}_{\text{pred-image}}$. In addition, we add a proposal loss to utilize the dynamic instance proposals for guiding the learning, $\mathcal{L}_{\text{proposal}}=\big\Arrowvert \sum_{i} (D_{:,:,i}^{t} - P_{:,:,i}^{t}) \big\Arrowvert_2^2$, where $P$ denotes the dynamic instance region proposals provided by the level of dynamic instance segmentation. Therefore, the total loss of the dynamics learning level is given by,
$
\mathcal{L}_{\text{DL}}=\mathcal{L}_{\text{pred-object}} + \lambda_{1}\mathcal{L}_{\text{pred-image}}+ \lambda_{2}\mathcal{L}_{\text{proposal}}.
$

\section{Experiments}
We compare MAOP with state-of-the-art action-conditioned dynamics learning baselines, AC Model \citep{oh2015action}, CDNA \citep{finn2016unsupervised}, and OODP \citep{zhu2018object}. AC Model adopts an encoder-LSTM-decoder structure, which performs transformations in hidden space and constructs pixel predictions. CDNA explicitly models pixel motions to achieve invariance to appearance. OODP trys to simultaneously learn object-based representations, relations and motion effects.
MAOP adopts a multi-level abstraction framework to decompose raw images into objects and predict instance-level dynamics based on actions and object relations.
OODP and MAOP both aim at learning object-level dynamics through an object-oriented learning paradigm, which decomposes raw images into objects and perform prediction based on object-level relations. OODP is only designed for class-level dynamics, while MAOP is able to learn instance-level dynamics. See Appendix for more implementation details.

\subsection{Generalization Ability and Sample Efficiency}
\begin{figure}[ht]

	\begin{center}
		\centering
		\includegraphics[width=0.9\columnwidth]{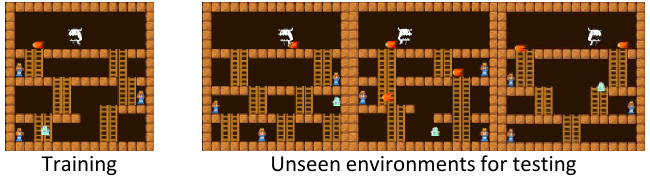}

		\caption{An Example of 1-to-3 zero-shot generalization. }
		\label{game}
	\end{center}

\end{figure}

We first evaluate zero-shot generalization and sample efficiency on \emph{Monster Kong} from Pygame Learning Environment \citep{tasfi2016PLE}, which allows us to test generalization ability over various scenes with different layouts. It is the advanced version of that used by \citet{zhu2018object}, which has a more general and complex setting. The monster wanders around and breathes out fires randomly, and the fires also move with some randomness. The agent randomly explores with actions \emph{up}, \emph{down}, \emph{left}, \emph{right}, \emph{jump} and \emph{noop}. All these dynamic objects interact with the environment and other objects according to the underlying physics engine. Moreover, gravity and jump model has a long-term dynamics effects, leading to a partial observation problem. To test whether our model can truly learn the underlying physical mechanism behind the visual observations and perform relational reasoning, we set the $k$-to-$m$ zero-shot generalization experiment (Figure \ref{game}), where we use $k$ different environments for training and $m$ different unseen environments for testing.

\begin{table*}[htp]

	\caption{Prediction performance on \emph{Monster Kong}. $k$-$m$ means the $k$-to-$m$ generalization problem. $\S$ indicates training with only $3000$ samples. ALL represents all dynamic objects. The first column shows the number of samples used for training the models.}
	\smallskip
	\centering
	\resizebox{2\columnwidth}{!}{
	\begin{tabular}{c @{~~} c @{~~} c@{~}c | c@{~}c | c@{~}c | c@{~}c | c@{~}c | c@{~}c | c@{~}c | c@{~}c  }
		\toprule
		&\multirow{3}*{Models}& \multicolumn{8}{c|}{Training environments} & \multicolumn{8}{c}{Unseen environments}\\
		\cmidrule(r){3-18}
		& & \multicolumn{2}{c}{1-5$^\S$} & \multicolumn{2}{c}{1-5} & \multicolumn{2}{c}{2-5} & \multicolumn{2}{c|}{3-5} & \multicolumn{2}{c}{1-5$^\S$} & \multicolumn{2}{c}{1-5} & \multicolumn{2}{c}{2-5} & \multicolumn{2}{c}{3-5}\\
		\cmidrule(r){3-18}
		& & Agent & All  & Agent & All & Agent & All & Agent & All & Agent & All & Agent & All & Agent & All & Agent & All \\
		\midrule
		& MAOP & \multicolumn{2}{c|}{3k} & \multicolumn{2}{c|}{100k} & \multicolumn{2}{c|}{100k} & \multicolumn{2}{c|}{100k} & \multicolumn{2}{c|}{-}  & \multicolumn{2}{c|}{-}  & \multicolumn{2}{c|}{\textbf{-}} & \multicolumn{2}{c}{-} \\
		%\cline{2-7}
		Training&OODP &  \multicolumn{2}{c|}{3k} & \multicolumn{2}{c|}{200k} & \multicolumn{2}{c|}{200k} & \multicolumn{2}{c|}{200k} & \multicolumn{2}{c|}{-}  & \multicolumn{2}{c|}{-}  & \multicolumn{2}{c|}{-} & \multicolumn{2}{c}{-} \\
		%\cline{2-7}
		Samples &AC Model &  \multicolumn{2}{c|}{3k} & \multicolumn{2}{c|}{500k} & \multicolumn{2}{c|}{500k} & \multicolumn{2}{c|}{500k} & \multicolumn{2}{c|}{-}  & \multicolumn{2}{c|}{-}  &  \multicolumn{2}{c|}{-} & \multicolumn{2}{c}{-} \\
		%\cline{2-7}
		&CDNA &  \multicolumn{2}{c|}{3k} & \multicolumn{2}{c|}{300k} & \multicolumn{2}{c|}{300k} & \multicolumn{2}{c|}{300k} & \multicolumn{2}{c|}{-}  & \multicolumn{2}{c|}{-}  & \multicolumn{2}{c|}{-} & \multicolumn{2}{c}{-} \\
		\midrule
		& MAOP  & 0.95 & 0.92 &  0.98 & 0.95 & 0.99 & 0.96 & 0.99 & 0.95 & \textbf{0.94} & \textbf{0.90} & \textbf{0.97} & \textbf{0.92} & \textbf{0.98} & \textbf{0.93} & \textbf{0.99} & \textbf{0.94}\\
		%\cline{2-7}
		0-error&OODP & 0.15 & 0.15 & 0.18 & 0.16 & 0.22 & 0.17 & 0.26 & 0.20 & 0.14 & 0.15 & 0.20 & 0.15 & 0.18 & 0.15 & 0.26 & 0.18\\
		%\cline{2-7}
		accuracy&AC Model & 0.01 & 0.36 & 0.87 & 0.94 & 0.80 & 0.93 & 0.77 & 0.92 & 0.01 & 0.20 & 0.08 & 0.16 & 0.30 & 0.48 & 0.45 & 0.66\\
		%\cline{2-7}
		&CDNA & 0.28 & 0.62 & 0.77 & 0.84 & 0.78 & 0.82 & 0.78 & 0.84 & 0.26 & 0.44 & 0.79 & 0.80 & 0.78 & 0.78 & 0.81 & 0.83 \\
		\midrule
		& MAOP & \multicolumn{2}{c|}{24.58} & \multicolumn{2}{c|}{21.96} & \multicolumn{2}{c|}{21.97} & \multicolumn{2}{c|}{23.04} & \multicolumn{2}{c|}{\textbf{29.67}} & \multicolumn{2}{c|}{\textbf{27.22}} & \multicolumn{2}{c|}{\textbf{25.55}} & \multicolumn{2}{c}{\textbf{24.30}}\\
		%\cline{2-7}
		Object &OODP & \multicolumn{2}{c|}{65.63} & \multicolumn{2}{c|}{66.44} & \multicolumn{2}{c|}{66.66} & \multicolumn{2}{c|}{64.73} & \multicolumn{2}{c|}{65.46} & \multicolumn{2}{c|}{67.41} & \multicolumn{2}{c|}{67.78} & \multicolumn{2}{c}{64.95}\\
		%\cline{2-7}
		RMSE &AC Model & \multicolumn{2}{c|}{71.02} & \multicolumn{2}{c|}{18.88} & \multicolumn{2}{c|}{22.39} & \multicolumn{2}{c|}{21.30} & \multicolumn{2}{c|}{77.24} & \multicolumn{2}{c|}{57.41} & \multicolumn{2}{c|}{55.45} & \multicolumn{2}{c}{43.48}\\
		%\cline{2-7}
		&CDNA & \multicolumn{2}{c|}{40.92} & \multicolumn{2}{c|}{24.52} & \multicolumn{2}{c|}{24.37} & \multicolumn{2}{c|}{24.18} & \multicolumn{2}{c|}{51.08} & \multicolumn{2}{c|}{37.15} & \multicolumn{2}{c|}{27.67} & \multicolumn{2}{c}{25.33}\\
		\bottomrule
	\end{tabular}
	}
	\label{tab:mk}

\end{table*}

To make a sufficient comparison with previous methods on object dynamics learning and video prediction, we conduct 1-5, 2-5 and 3-5 generalization experiments with a variety of evaluation indices. We use $n$-error accuracy to measure the performance of object dynamics prediction, which is defined as the proportion that the difference between the predicted and ground-true agent locations is less than $n$ pixel. We also add an extra pixel-based measurement (denoted by object RMSE), which compares the pixel difference near dynamic objects between the predicted and ground-truth images.

\begin{figure}[ht]

	\begin{center}
		\includegraphics[width=0.98\columnwidth]{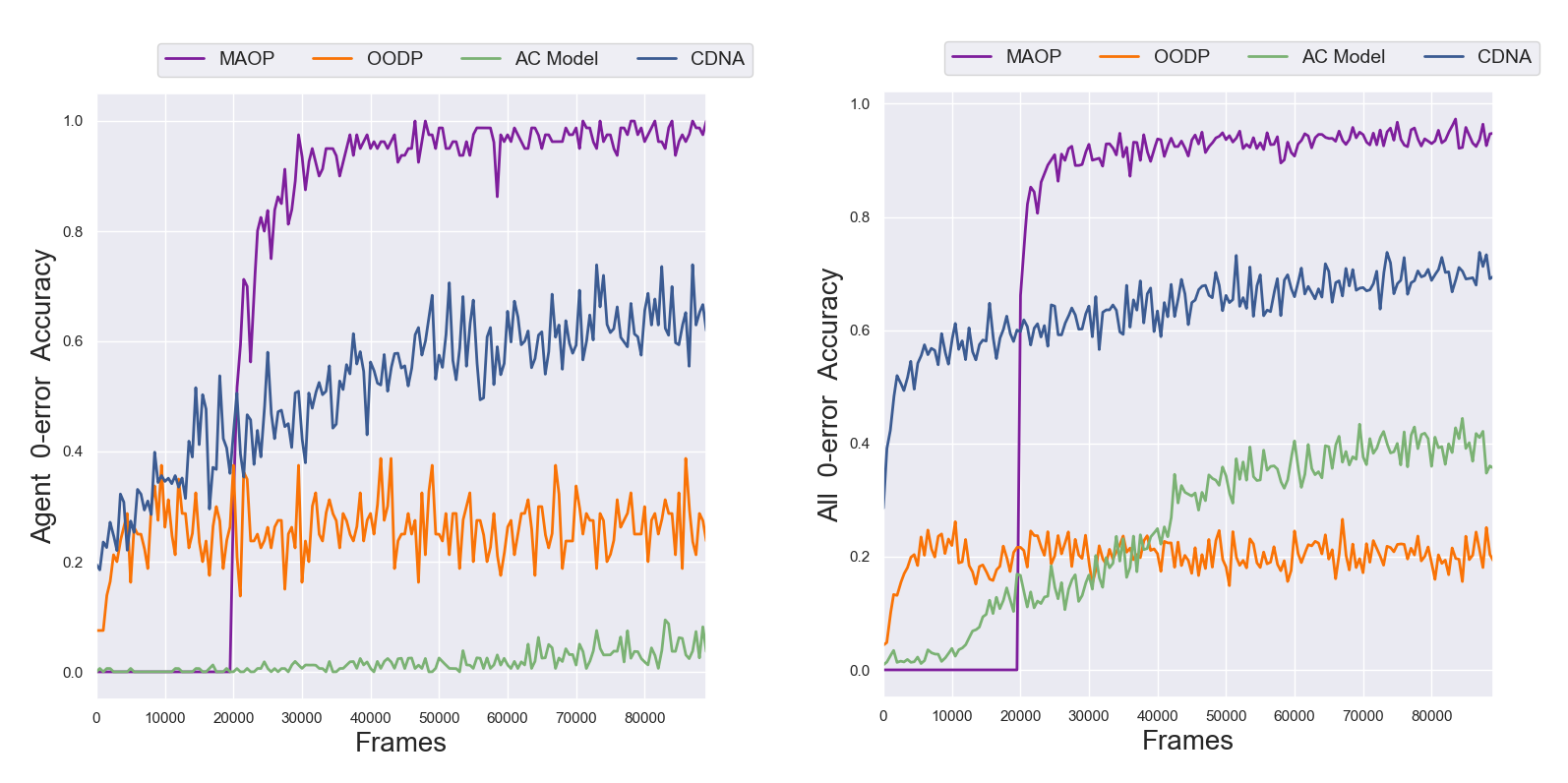}

		\caption{The performance of object dynamics prediction in unseen environments as training progresses on 3-to-5 generalization problem of \emph{Monster Kong}. Since we use the first 20k samples to train the level of dynamic instance segmentation, the curve of MAOP starts at iteration 20001.}
		\label{a0}
	\end{center}

\end{figure}

As shown in Table \ref{tab:mk}, MAOP significantly outperforms other methods in all experiment settings in terms of generalization ability and sample efficiency of object dynamics learning. It can achieve 90\% 0-error accuracy in unseen environments even trained with 3k samples from a single environment, while other methods have a much lower accuracy (less than 45\%). In addition, MAOP with only 3k training samples outperforms CDNA using 300k samples. Although AC Model achieves high accuracy in training environments, its performance in unseen scenes is much worse, which is probably because its pure pixel-level inference easily leads to overfitting. CDNA performs better than AC Model, but still cannot efficiently generalize with limited training samples. Since OODP has structural limitation and optimization difficulty, it has innate difficulty on frames with multiple dynamic objects. In Figure \ref{a0} and Figure \ref{ax}, we also plot the learning curve of these models. Compared to other models, MAOP achieves higher prediction accuracy for unseen environments at a faster rate during the training process. We further add a video (see \url{https://drive.google.com/drive/folders/1vI8SI16UiGknnySSwtSpCzX2AeLbmbdw?usp=sharing}) for better perceptual understanding of prediction performance in unseen environments.

\begin{table}[htp]

	\caption{Accuracy of dynamics prediction on \emph{Flappy Bird} and \emph{Freeway}. Since only the agent's ground-true location is accessible in ALE, 
		we just show the dynamics prediction of the agent. Actually, we observe that predictions of other dynamic objects are also accurate by comparing predicted with ground-true images (see Figure \ref{fw} in Appendix).}
	\smallskip
	\resizebox{1\columnwidth}{!}{
	
	\begin{tabular}{ c @{~} c@{~} c | c @{~}c  |  c @{~}c @{~}c }
		\toprule
		\multirow{3}*{Models}& \multicolumn{4}{c}{Flappy Bird (0-acc)} & \multicolumn{3}{c}{Freeway (Agent)}\\
		\cmidrule(r){2-8}
		& \multicolumn{2}{c}{100 samples} & \multicolumn{2}{c}{300 samples} & \multicolumn{3}{c}{100 samples} \\
		\cmidrule(r){2-8}
		& Agent & All & Agent & All & 0-acc & 1-acc & 2-acc\\
		\midrule
		MAOP & \textbf{0.83} & \textbf{0.89} & \textbf{0.83} & \textbf{0.92} &  \textbf{0.78} & \textbf{0.84} & \textbf{0.89}  \\
		
		OODP & 0.01 & 0.18 & 0.02 & 0.15 & 0.26 & 0.33 & 0.42 \\
		%\cline{2-7}
		AC Model & 0.03 & 0.18 & 0.04 & 0.23 & 0.31 & 0.38 & 0.42 \\
		%\cline{2-7}
		CDNA & 0.13 & 0.77 & 0.30 & 0.81 &  0.42 & 0.43 & 0.47 \\
		\bottomrule
	\end{tabular}
}
	\centering

	\label{ff}
\end{table}

We also evaluate MAOP on \emph{Flappy Bird} and \emph{Freeway}. \emph{Flappy Bird} is a side-scroller game with a moving camera. \emph{Freeway} is an Atari game, which has a large number of dynamic objects. Since the testing environments will be similar with the training ones without limitation of samples, we limit the training samples to form a sufficiently challenging generalization task. MAOP still outperforms existing baseline methods (Table \ref{ff}), which demonstrates that MAOP is effective for the concurrent dynamics prediction of a large number of objects. In addition, we conduct a modular test to better understand the contribution of each learning level (see Modular Test and Ablation Study in Appendix). The results show that each level of MAOP can independently perform well and has a good robustness to the proposals generated by the more abstracted level. Taken together, the above results demonstrates that MAOP has superiority of sample efficiency and generalization ability, which suggests MAOP is good at relational reasoning and learns the object-level dynamics, rather than learn some patterns from mass data to recover the dynamics as the conventional neural networks do.

To further test the priority and limitation of MAOP, we have applied MAOP on a diverse set of games on Atari. Testing results with 3k training samples on \emph{Skiing}, \emph{MsPacman}, \emph{Krull}, \emph{Pong}, \emph{MontezumaRevenge}, and \emph{Breakout} are shown in Table \ref{atari}. We observe a superior performance of MAOP over baseline models on \emph{Skiing}, \emph{MsPacman}, \emph{Krull} and \emph{Pong}, and a slightly worse performance on \emph{Breakout}. Because our model is designed for scenes with multiple dynamic objects, its performance may be lower than some simpler baseline methods on environments with only one or two dynamic objects, such as \emph{MontezumaRevenge} and \emph{Breakout}. 

%In addition, a very long-range history is required for predicting the future trajectory of the bouncing ball in \emph{Breakout} thus the good performance of CDNA may benefit from the LSTM architecture that is good at long-range prediction. 

\begin{table*}[t]
	\centering
	\caption{Accuracy of dynamics prediction on six Atari games. }\label{atari}
	\smallskip
	\resizebox{2\columnwidth}{!}{
		\begin{tabular}{c  c c c @{~}| c c c @{~}|c c c @{~}| c c c @{~}| c c c @{~}| c c c @{~}}
			\toprule
			\multirow{2}*{Model}& \multicolumn{3}{c}{Skiing} & \multicolumn{3}{c}{MsPacman} & \multicolumn{3}{c}{Krull} & \multicolumn{3}{c}{Pong} & \multicolumn{3}{c}{MontezumaRevenge} & \multicolumn{3}{c}{Breakout}\\
			\cmidrule(r){2-19}
			& 0acc & 1acc & 2acc & 0acc & 1acc & 2acc & 0acc & 1acc & 2acc & 0acc & 1acc & 2acc & 0acc & 1acc & 2acc & 0acc & 1acc & 2acc\\
			\midrule
			MAOP  & \textbf{0.97} & \textbf{0.99} & \textbf{1.00} & \textbf{0.65} & \textbf{0.88} & \textbf{0.94} & 0.14 & \textbf{0.48} & \textbf{0.73} & 0.63 & \textbf{0.73} & \textbf{0.83} & 0.95 & \textbf{1.00} & \textbf{1.00} & 0.52 & 0.66 & 0.77 \\
			OODP  & 0.62 & 0.80 & 0.90 & 0.30 & 0.35 & 0.46 & 0.02 & 0.08 & 0.16 &  0.46 & 0.64 & 0.66 & 0.66 & 0.92 & 0.99 & 0.73 & 0.66 & \textbf{0.80} \\
			AC Model & 0.27 & 0.47 & 0.56 & 0.44 & 0.52 & 0.54 & 0.01 & 0.05 & 0.13 & 0.37 & 0.40 & 0.42 & 0.63 & 0.79 & 0.95 & 0.45 & 0.57 & 0.66 \\
			CDNA & 0.76 & 0.95 & 0.99 & 0.52 & 0.68 & 0.74 & \textbf{0.27} & 0.41 & 0.51 & \textbf{0.65} & 0.66 & 0.79 & \textbf{0.96} & 1.00 & 1.00 & \textbf{0.63} & \textbf{0.71} & 0.77 \\
			\bottomrule
		\end{tabular}
	}
\end{table*}

\subsection{Model-Based Planning in Unseen Environments}
Although RL has achieved considerable successes, most RL researches tend to ``train on the test set" \cite{nichol2018gotta,reRL}. It is critical yet challenging to develop model-based RL approaches that support generalization over unseen environments. Monte Carlo tree search (MCTS) \cite{browne2012survey} is developed to leverage the environment models to conduct efficient lookahead search, which has shown remarkable effectiveness on long-term planning, such as AlphaGo \cite{silver2016mastering}. Considering that our learned dynamics model can efficiently generalize to unseen environments, we can directly use our learned model to perform MCTS in unseen environments. To perform long-range planning, we first test our performance of long-range prediction, as shown in Table \ref{long} in Appendix. MAOP only trained for 1-step prediction can achieve 90\% 2-error accuracy in unseen environments when predicting 3 steps of the future, while the accuracy is 73\% when predicting 6 steps of the future, which is also a satisfactory performance for lookahead search. Figure \ref{8p} in Appendix illustrates a case visualizing the 6-step prediction of MAOP in unseen environments.

We evaluate our performance of model-based planning on \emph{Monster Kong}. In this game, the goal of the agent is to approach the princess and a reward will be given when the straight-line distance from agent to princess gets smaller than that in the agent's history. The value of such a reward is proportional to the shrinking distance. The agent will win with an extra reward +5 when touching the princess, and lose with an extra reward -5 when hitting the fires. To gain a better understanding of the contribution of MAOP to the MCTS agent, we compare MCTS in conjunction with MAOP to DQN \cite{mnih2015human} and to an ablation (i.e., using the real simulator of the unseen environments in MCTS). We provide the same ground-true reward functions for all dynamics model during MCTS. We conduct random experiments in 5 unseen environments, where the agent and the princess randomly generate. We train all models in training environments with 5k samples, and test zero-shot generalization of the model-free behavior policy (i.e., DQN) and model-based planning policy (i.e., MAOP-based, CDNA-based, OODP-based and AC-based) in unseen environments. 

As shown in Table \ref{mcts}, MAOP achieves almost the same performance with the true environment model for model-based planning in unseen environments and significantly outperforms other baseline models and DQN. The model-free approach DQN tends to overfit the training environments and cannot learn to plan in unseen environments, leading to a much higher death rate and a much lower score. The learning curves in Figure \ref{mctsc} in Appendix also verify this. In addition, we observe that MCTS in conjunction with MAOP acquires intriguing forward-looking skills, such as jumping over the fires and jumping across the big gap that are critical for survival and reaching the goal (we provide videos for the learned policies at \url{https://drive.google.com/drive/folders/1vI8SI16UiGknnySSwtSpCzX2AeLbmbdw?usp=sharing}).

\begin{table}[htp]
	\centering
	\caption{The performance of using MCTS with different dynamics models, and DQN in unseen environments. REAL indicates the real simulator. Time Out indicates exceeding 100 steps. Reward is averaged over 21 runs.}
	\smallskip
	\begin{tabular}{c  c @{~} c @{~} c @{~} c }
		\toprule
		Methods& Reward  & Win & Lose & Time Out \\
		\midrule
		MCTS + MAOP  & 38.19 & 47.62\% & 9.52\%  & 42.86\% \\
		MCTS + REAL &  38.41 & 52.38\% & 9.52\%  & 38.10\% \\
		MCTS + CDNA & 6.83 & 0\% & 33.33\% & 66.67\%  \\
		MCTS + OODP	& 13.95 & 0\% & 52.38\% & 47.62\%  \\
		MCTS + AC & 7.50 & 0\% & 47.62\% & 52.38\%  \\
		DQN & 13.67 & 26.7\% & 23.8\% & 49.5\% \\
		\bottomrule
	\end{tabular}
	
	\label{mcts}
\end{table}

\subsection{Interpretable Representations and Knowledge}
MAOP takes a step towards interpretable dynamics model learning. Through interacting with environments, it learns visually and semantically interpretable knowledge in a self-supervised manner, which contributes to unlocking the ``black box'' of the dynamics prediction and potentially opens the avenue for further researches on object-oriented RL, model-based RL, and hierarchical RL.

\begin{figure}[ht]

	\begin{center}
		\includegraphics[width=\columnwidth]{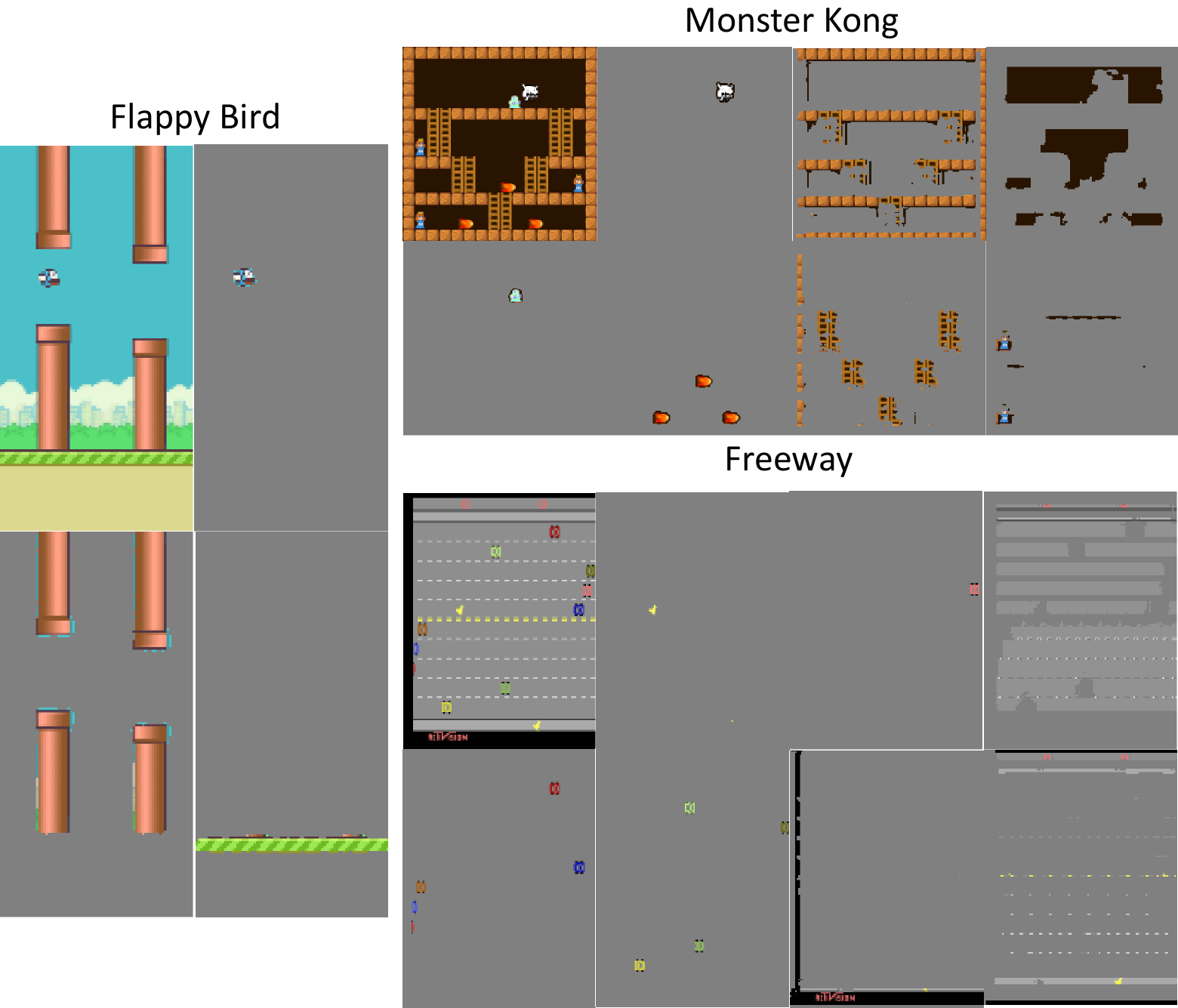}

		\caption{Visualization of the masked images in unseen environments. Top left corner is the raw image.}
		\label{om}
		
	\end{center}

\end{figure}

\textbf{Visual Interpretability.}
To demonstrate the model interpretability of MAOP in unseen environments, we visualize the learned masks of dynamic and static objects. We highlight the attentions of object masks by multiplying the raw images by the binarized masks. Note that MAOP does not require the actual number of objects but a maximum number and some learned object masks may be redundant. Thus, we only show the informative object masks. As shown in Figure \ref{om}, our model captures all the key objects in the environments including the controllable agents (cowboy, bird, and chicken), the uncontrollable dynamic objects (monster, fires, pipes and cars), and the static objects that have effects on the motions of dynamic objects (ladders, walls and the free space), which demonstrates that model can learn disentangled object representations and distinguish the objects by both appearance and dynamic property.

%\textbf{Semantic interpretability.}
%Since we use a disentangle dynamics learning paradigm, we can easily decouple and decode our learned knowledge including the object-level effects from actions, pairwise object-to-object relations and historical states. Here, we elaborate two examples (Figure ).

\textbf{Dynamical Interpretability.}
To show the dynamical interpretability behind image prediction, we test our predicted motions by comparing RMSEs between the predicted and ground-truth motions in unseen environments (Table \ref{tab:motion} in Appendix). Intriguingly, most predicted motions are quite accurate, with the RMSEs less than 1 pixel. Such a visually indistinguishable error also verifies the accuracy of our dynamics learning.

\textbf{Discovery of the Controllable Agent.}
With the learned knowledge in MAOP, we can easily uncover the action-controlled agent from all the dynamic objects, which is useful semantic information that can be used in heuristic algorithms. For example, it allows allows agents to efficiently explore (e.g., contingency awareness \citep{choi2019contingency}, empowerment \citep{karl2017unsupervised}, megalomania-drivenness \citep{song2019mega}, and distance-based rewards \citep{srinivas2018universal}). Specifically, the object that has the maximal variance of total effects over actions is the action-controlled agent. Denote the total effects as $E^{t}_i=(\sum_{j} E^{t}(c_{i},j))+ E^{t}_{\text{self}}(c_{i})$, the label of the action-controlled agent is calculated as, ${arg\,max}_i \sum_t Var_{a^{t}}(E^{t}_i)$. We observe that our discovery of the controllable agent achieves right or near $100\%$ accuracy in unseen environments (see Table \ref{control} in Appendix).
%The histogram in Figure \ref{control} plotting the ground-truth label distribution of our discovered action-controlled agents clearly demonstrates that our discovery of the controllable agent achieves right or near $100\%$ accuracy.
\section{Conclusion and Discussion}
This paper presents a self-supervised multi-level learning framework for learning action-conditioned object-based dynamics. It enables sample-efficient and interpretable model learning, and achieves zero-shot generalization over novel environments with multiple dynamic objects and different static object layouts.
% that have multiple controllable and uncontrollable dynamic objects and different static object layouts
The learned dynamics model enables an agent to directly plan in unseen environments. MAOP can easily generalize the learned knowledge over environments with similar objects but may not work well with those with totally new objects, which is an important direction for future work.

As abrupt changes (e.g., colors) are often predictable from a long-term view or memory, our model can be extended to more domains by incorporating memory networks (e.g., LSTM). In addition, our future work includes extending our model for deformation prediction (e.g., object appearing, disappearing and non-rigid deformation) and incorporating a camera motion prediction network module introduced by \cite{vijayanarasimhan2017sfm} for applications such as FPS games and autonomous driving. Learning 3D dynamics from 2D video is extremely challenging. Conventional neural networks try to learn such 3D dynamics by remembering some patterns from 2D data as they do for the non-rigid deformation, such as AC Model \citep{oh2015action} and CDNA \citep{finn2016unsupervised}. This approach achieves good performance in training environments, but it requires a large number of data and does not really recover the true 3D dynamics model. To learn generalized 3D dynamics model, object-oriented learning paradigm in conjunction with 3D CNN (3D data input) is necessary, which is an important direction for future work.

\bibliography{AAAI-ZhuG.4599}
\bibliographystyle{aaai}

\clearpage
\section{Appendix}
\setcounter{table}{0}
\setcounter{figure}{0}
\setcounter{algorithm}{0}
\renewcommand\thetable{S\arabic{table}}
\renewcommand\thefigure{S\arabic{figure}}
\renewcommand\thealgorithm{S\arabic{algorithm}}

\subsection{Object-Oriented Dynamics Learning Paradigm}
Algorithm \ref{paradigm} illustrates the learning paradigm of object based dynamics and the interactions of its components (Section Object-Oriented Dynamics Learning Level in the main body).

\subsection{Instance Localization}\label{il}
Instance localization is a common technique in context of supervised region-based object detection  \citep{girshick2014rich,girshick2015fast,ren2015faster,he2017mask,liu2018deep}, which localizes objects on raw images with regression between the predicted bounding box and the ground truth. Here, we propose an unsupervised approach to perform dynamic instance localization on dynamic object masks learned by Object Detector. Our objective is to sample a number of region proposals on the dynamic object masks and then select the regions, each of which has exactly one dynamic instance. In the rest of this section, we will describe these two steps in detail.

%The first part generates region proposals which define the set of candidate regions for the following instance segmentation. The second part is Dynamic Instance Segmentation Network that leverages object appearance and dynamics consistency to derive the instance segmentation. The third part performs instance mask selection to eliminate redundant and noisy instance masks, which is similar to non-maximum suppression (NMS) \citep{ren2015faster}.

\textbf{Region proposal sampling.}
We design a learning-free sampling algorithm for sampling region proposals on object masks. This algorithm generates multi-scale region proposals with a full coverage over the input mask. Actually, we adopt multi-fold full coverage to ensure that pixels of the potential instances are covered at each scale.
The detailed algorithm is described in Algorithm \ref{RPS}.

\textbf{Instance mask selection.}
Instance mask selection aims at selecting the regions, each of which contains exactly one dynamic instance, based on the discrepancy loss $\mathcal{L}_{\text{instance}}$ (Section Dynamic Instance Segmentation Level in the main body). To screen out high-consistency, non-overlapping and non-empty instance masks at the same time, we integrate Non-Maximum Suppression (NMS) and Selective Search (SS) \citep{uijlings2013selective} in the context of region-based object detection \citep{girshick2014rich,girshick2015fast,ren2015faster,he2017mask,liu2018deep} into our algorithm. We select the top $K$ dynamic instances with the max scores, where $K$ is the maximum number of dynamic instances. If $K$ is larger than the real number, some selected instance masks will be blank, which does not affect the prediction performance.

\subsection{Modular Test and Ablation Study}
\begin{figure*}[htbp]
	\centering
	\includegraphics[width=1.9\columnwidth]{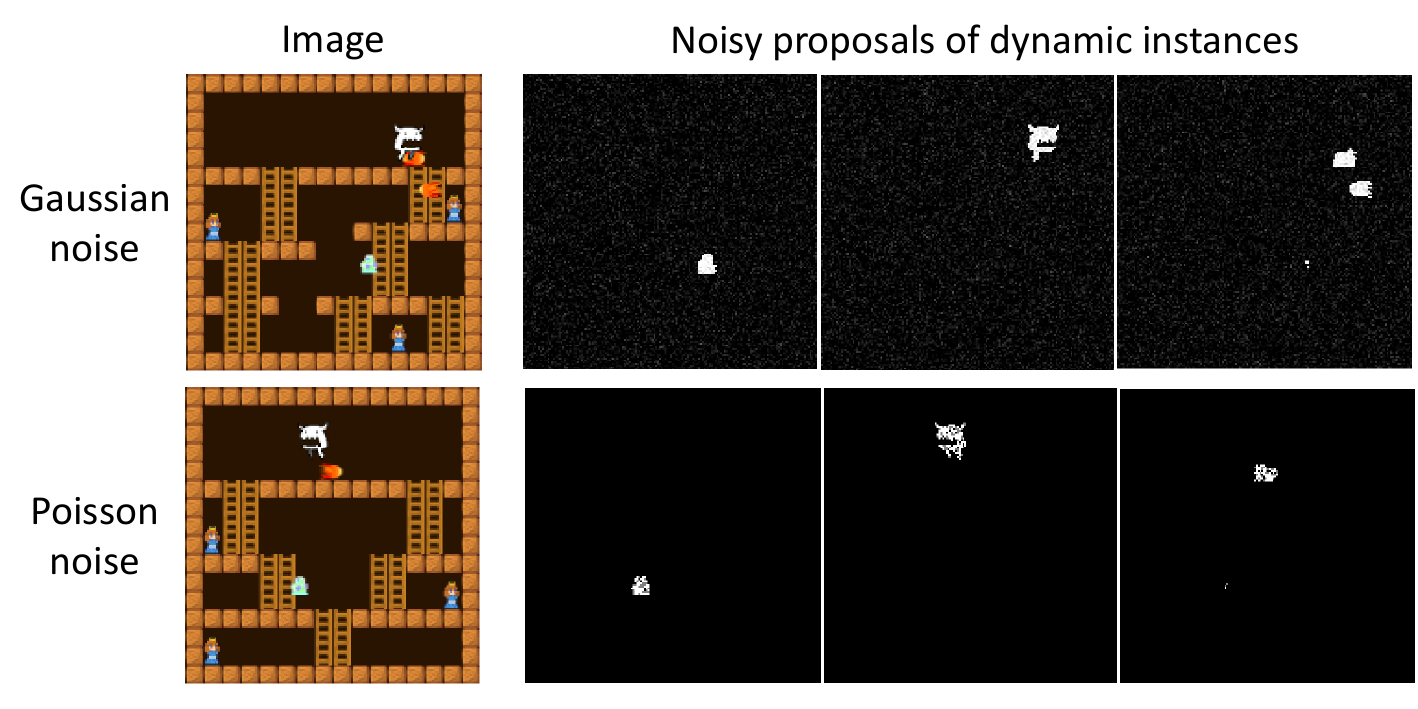}
	\caption{Noisy region proposals of dynamic instances. Zoom in to see the details.}
	\label{noise}
	
\end{figure*}

We conduct a modular test to better understand the contribution of each learning level. First, we investigate whether the level of dynamics learning can learn the accurate dynamics model when coarse region proposals of dynamic instances are given. We remove the other two levels and replace them by the artificially synthesized coarse proposals of dynamic instances to test the independent performance of the dynamics learning level. Specifically, the synthesized data are generated by adding standard Gaussian or Poisson noise on ground-true dynamic instance masks (Figure \ref{noise}). As expected, the level of dynamics learning can learn accurate dynamics of all dynamic objects given coarse proposals of dynamic instances (Table \ref{noiseres}).

\begin{figure*}[htbp]
	\centering
	\includegraphics[width=1.9\columnwidth]{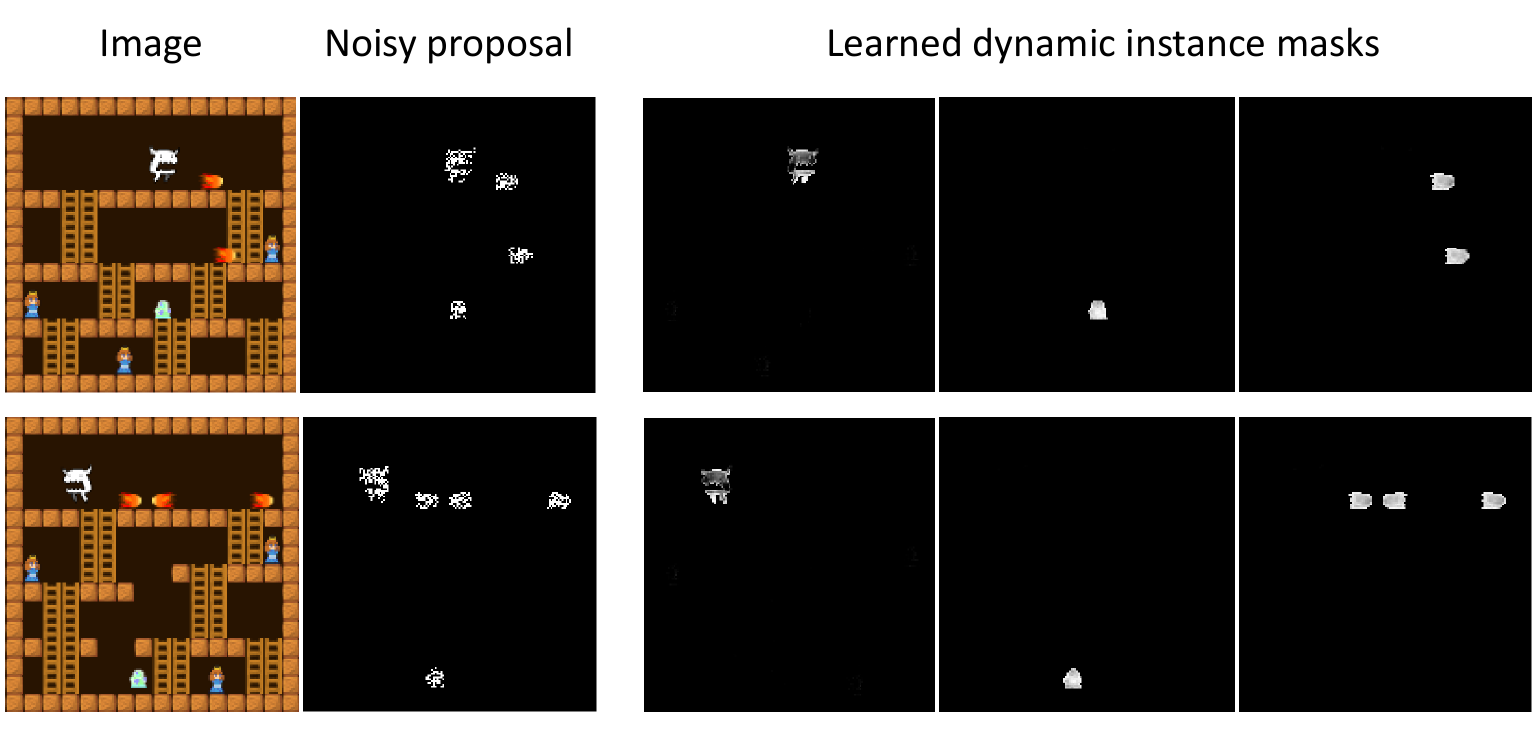}
	\caption{The learned dynamic instance masks in the level of dynamic instance segmentation with noisy foreground proposals. }
	\label{dis}
	
\end{figure*}

Similarly, we test the independent performance of the dynamics instance segmentation level. We replace the foreground proposal generated by the motion detection level with the artificially synthesized noisy foreground proposal. Figure \ref{dis} demonstrates our learned dynamic instances in the level of dynamic instance segmentation, which demonstrates the competence of the dynamic instance segmentation level. Taken together, the modular test and ablation study show that each level of MAOP can independently perform well and has a good robustness to the proposals generated by the more abstracted level.

\subsection{Robustness to mask number}
Note that our framework does not require the actual number of object classes, but needs to set a maximum number (usually 10 is enough). When they do not match, some learned object masks may be redundant, which does not affect the accuracy of predictions. We have conducted experiments to confirm this, as shown in Table \ref{mn}. 

\subsection{Supplementary Tables and Figures}\label{atf}
\begin{figure*}[ht]
	\begin{center}
		\centering
		\includegraphics[width=1.9\columnwidth]{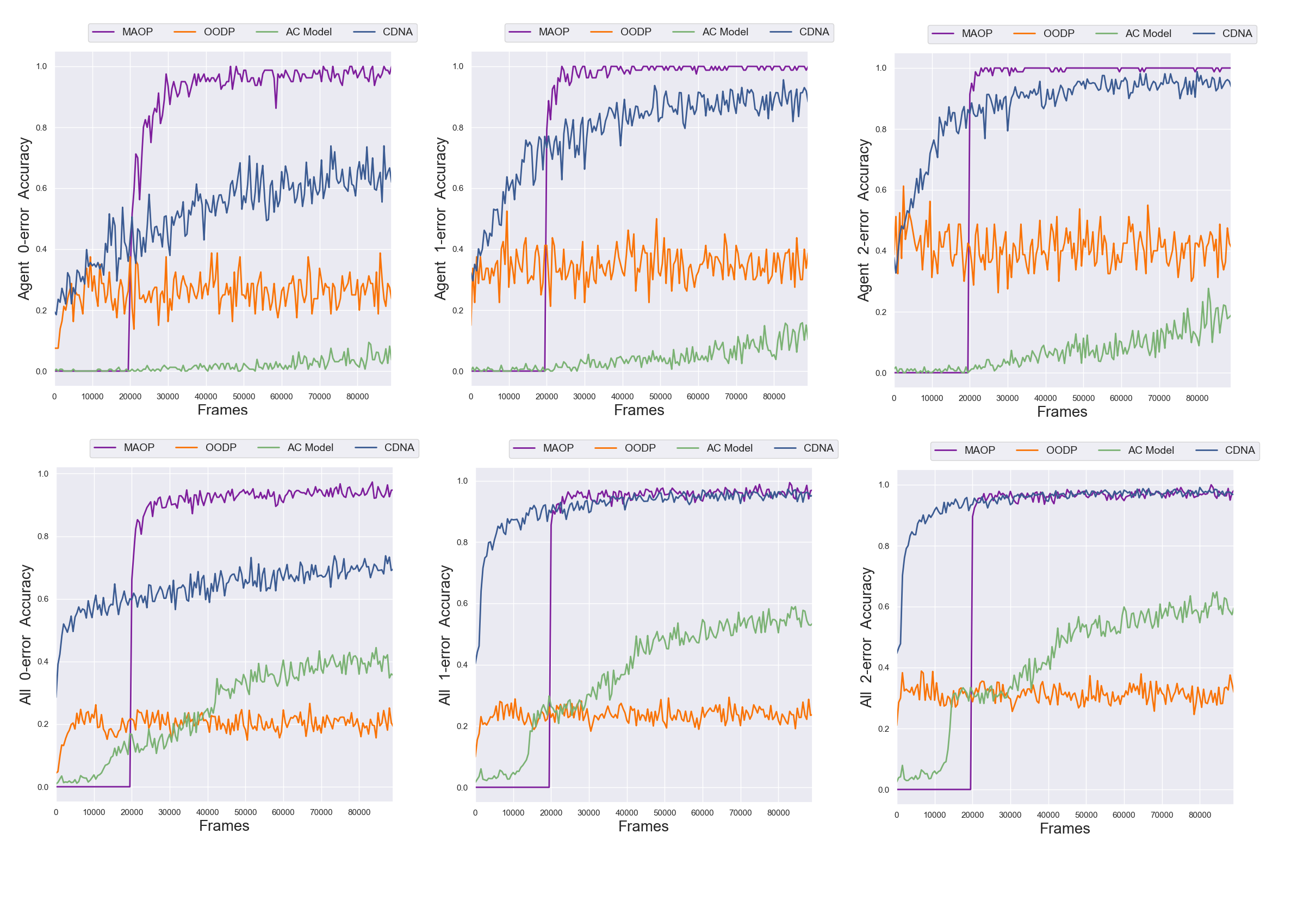}
		\caption{The performance of object dynamics prediction in unseen environments as training progresses on 3-to-5 generalization problem of \emph{Monster Kong}. ``Agent" represents the dynamics of the agent, while ``all" represents the dynamics of all dynamic objects. Since we use the first 20k samples to train the level of dynamic instance segmentation, the curve of MAOP starts at iteration 20001.}
		\label{ax}
	\end{center}
\end{figure*}

\begin{figure*}[htbp]
	\centering
	\includegraphics[width=1.9\columnwidth]{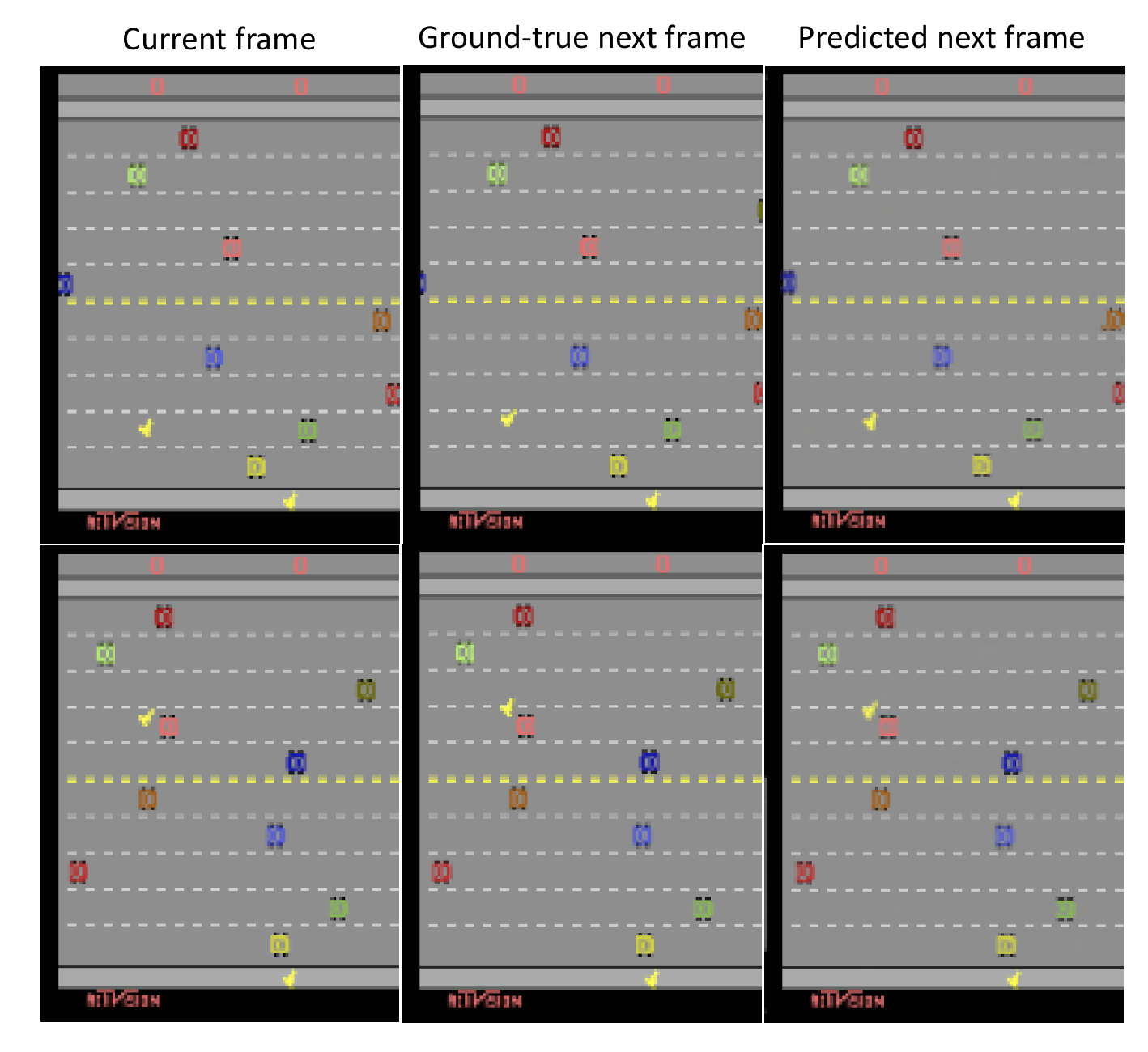}
	\caption{Image predictions in testing environments on \emph{Freeway}. Since just the agent's ground-true location is accessible in Arcade Learning Environment, we can only examine the predictions of other dynamic objects by comparing the predicted with ground-true images. These two samples are not cherry-picked. From the figure, we can observe that the errors between the predicted and ground-true images are visually indistinguishable, which suggests that our prediction of all dynamic objects are accurate.}
	\label{fw}
	
\end{figure*}

\begin{figure*}[ht]
	\begin{center}
		\centering
		\includegraphics[width=1.9\columnwidth]{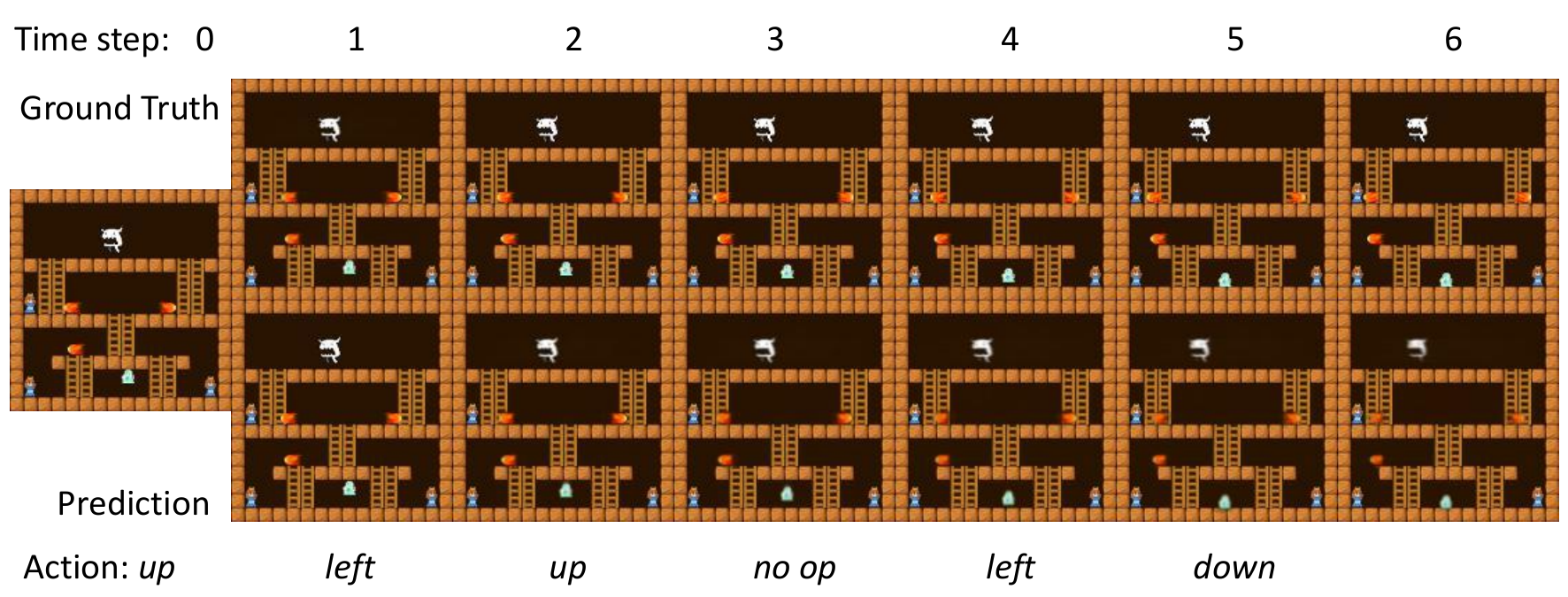}
		\caption{A case shows the 6-step prediction of our model in unseen environments on 3-to-5 generalization problem of \emph{Monster Kong}. }
		\label{8p}
	\end{center}
\end{figure*}

\begin{figure*}[ht]
	\begin{center}
		\centering
		\includegraphics[width=1.2\columnwidth]{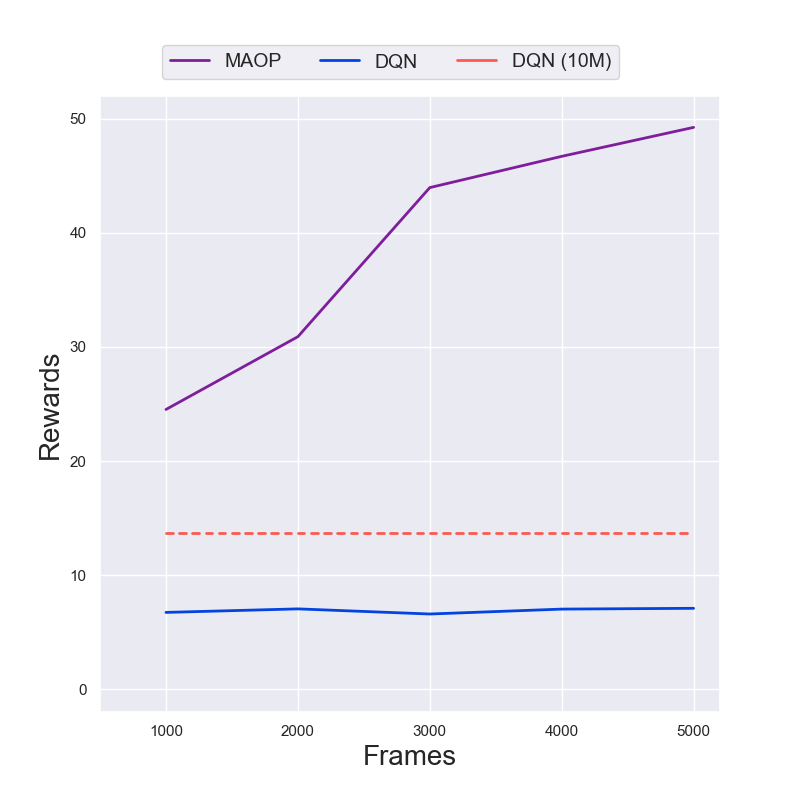}
		\caption{Performance of learned polices as training frames increase. }
		\label{mctsc}
	\end{center}
\end{figure*}

In addition to the above mentioned tables and figures, here we supplement the rest of supplementary tables and figures: \textbf{Table \ref{long}} (mentioned in Section Model-Based Planning in Unseen Environments of the main body), \textbf{Table \ref{tab:motion}} (mentioned in Interpretable Representations and Knowledge of the main body), \textbf{Table \ref{control}} (mentioned in Interpretable Representations and Knowledge of the main body), \textbf{Figure \ref{ax}} (mentioned in Section Generalization Ability and Sample Efficiency of the main body), \textbf{Figure \ref{fw}} (mentioned in Table 2 of the main body), \textbf{Figure \ref{8p}} (mentioned in Section Model-Based Planning in Unseen Environments of the main body), and \textbf{Figure \ref{mctsc}} (mentioned in Section Model-Based Planning in Unseen Environments of the main body).

\subsection{Implementation Details for Experiments}\label{implement}
\begin{figure}[htbp]
	\centering
	\includegraphics[width=\columnwidth]{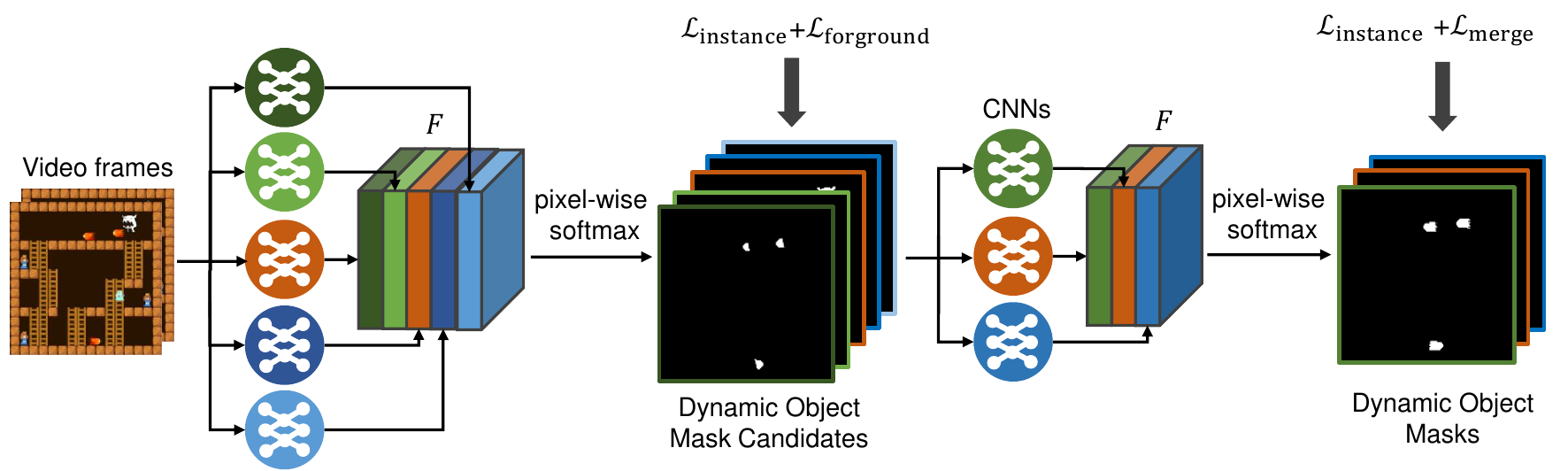}
	\caption{Architecture of the dynamic instance segmentation level, which consists of Instance Splitter and Merging Net.}
	\label{ISMN}
\end{figure}

\begin{figure}[htbp]
	\centering
	\includegraphics[width=\columnwidth]{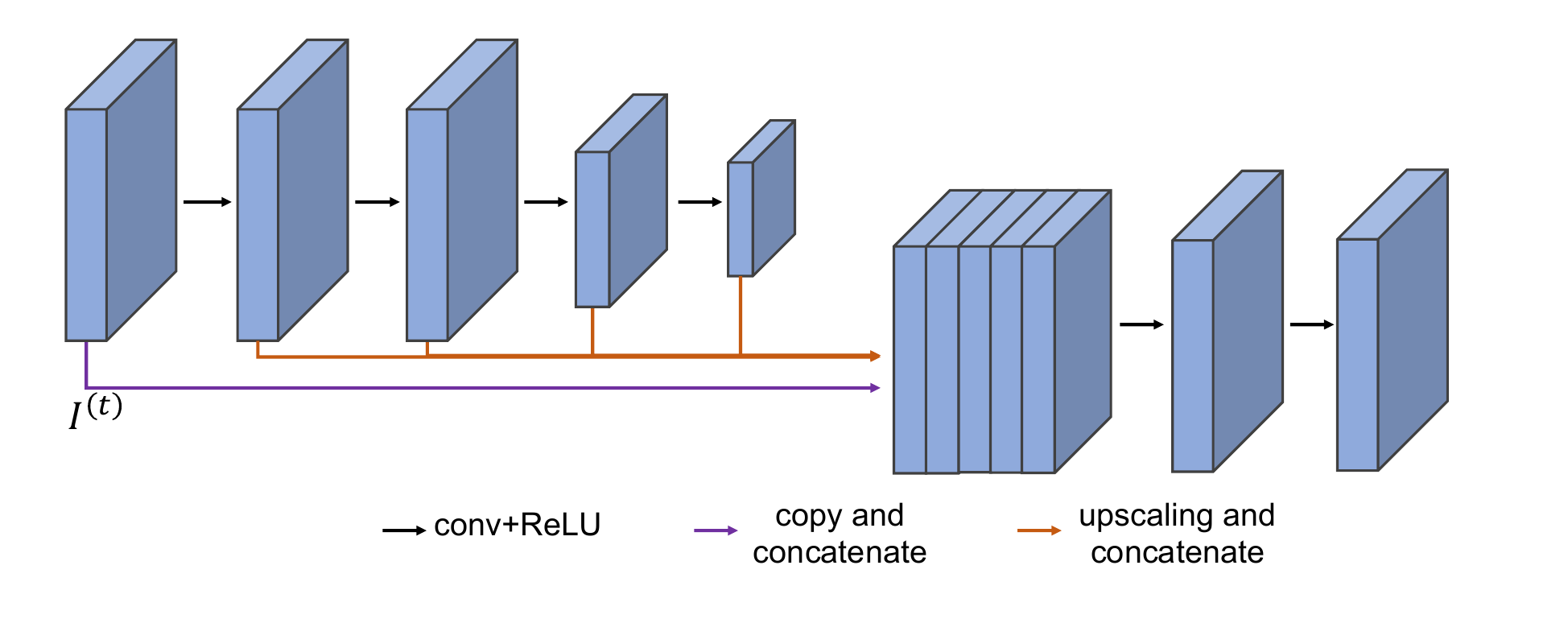}
	\caption{Architecture of the CNNs in Object Detector.}
	\label{ODstructure_1}
\end{figure}

We have tuned many  hyperparameters and network architectures and observed that our framework is not sensitive to most of them. Here we supplement one set of implementation settings. The neural network architecture of the dynamic instance segmentation level (consisting of Instance Splitter and Merging Net) is shown as Figure \ref{ISMN}. Object Detector in the dynamics learning level has the similar architecture with Instance Splitter. The CNNs in Object Detector are shown as Figure \ref{ODstructure_1}.

Denote $Conv(F,K,S)$ as the convolutional layer with the number of filters $F$, kernel size $K$ and stride $S$. Let $R(), S()$ and $BN()$ denote the ReLU layer, sigmoid layer and batch normalization layer \citep{ioffe2015batch}. The CNNs in Merging Net are connected in the order: $R(BN(Conv(32,1,1)))$, $R(BN(Conv(1,1,1)))$. The 6 convolutional layers (Figure \ref{ODstructure_1}) in Object Detector can be indicated as $R(BN(Conv(32,3,1)))$, $R(BN(Conv(32,3,1)))$, $R(BN(Conv(16,3,2)))$, $R(BN(Conv(16,3,2)))$, $R(BN(Conv(32,1,1)))$ and $BN(Conv(1,3,1))$, respectively. The CNNs in Relation Net are connected in the order: $R(BN(Conv(16,3,2)))$, $R(BN(Conv(16,3,2)))$, $R(BN(Conv(16,3,2)))$, and $R(BN(Conv(16,3,2)))$. The last convolutional layer is reshaped and fully connected by the 64-dimensional hidden layer and the 2-dimensional output layer successively. Inertia Net has the same architecture and hyperparameters as Relation Net.

Since Relation Nets model pairwise relations, the complexity grows quadratically with the total number of object classes. However, each Relation Net is very small (< 20k parameters) because it takes tailored masks (e.g., 32$\times$2) as input and has limited layers. In our experiments, MAOP has a much smaller number of parameters (about 1M) compared to previous models (about 40M in AC Model, 20M in CDNA, and 1M in OODP).

The detailed experimental settings and hyperparameters for training MAOP on \emph{Monster Kong}, \emph{Flappy Bird} and Atari games are listed as follows:
\begin{itemize}
	\item We use random exploration on \emph{Monster Kong}. We adopt an expert guided random exploration on \emph{Flappy Bird} and Atari games, because a totally random exploration will lead to an early death of the agent even at the very beginning. Although we use these exploration methods in our experiments, our framework can support smarter exploration strategies, such as curiosity-driven exploration \citep{pathak2017curiosity}.
	
	\item The weights of losses, $\lambda_{1}$, $\lambda_{2}$, $\lambda_{3}$, $\lambda_{4}$ are 100, 1, 10, and 10, respectively. All the $l_2$ losses are divided by $HW$ to keep invariance to the image size.
	
	\item The optimizer is Adam \citep{DBLP:journals/corr/KingmaB14} with learning rate $1 \times 10^{-3}$. Batchsize is 8.
	\item The raw images of \emph{Monster Kong}, \emph{Flappy Bird} and Atari games are resized to $160 \times 160 \times 3$, $160 \times 80 \times 3$, and $160 \times 120 \times 3$ , respectively.
	\item The maximum number of masks is 10 on \emph{Monster Kong}, 10 on \emph{Flappy Bird} and 20 on Atari games.
	\item The size of the horizon window $w$ is 33 on \emph{Monster Kong}, 41 on \emph{Flappy Bird}, and 33 on Atari games.
	\item To augment the interactions of instances when training Instance Splitter, we random sample two region proposals and combine them into a single region proposal with double size.
	
\end{itemize}

The detailed hyperparameters for running MCTS with MAOP, OODP, CDNA, AC Model, and real simulator on \emph{Monster Kong} are listed as follows:
\begin{itemize}
	\item The number of trajectories is 500.
	\item The maximum-depth of each trajectory is 6.
	\item The exploration parameter used in Upper Confidence Bounds for Trees (UCT) is 5.
	\item The number of rollouts in each simulation is 8.
	\item At the end of each search, the agent selects the action with maximum visit count.
\end{itemize}

\begin{algorithm*}[htbp]
	\caption{Object-Oriented Dynamics Learning.} \label{aloodl}
	\hspace*{0.02in} {\bf Input:}
	A sequence of video frames $I^{t-h:t}$ with length $h$, input action $a^t$ at time $t$.
	\begin{algorithmic}[1]
		\STATE Object masks $\O^{t-h:t} \gets \text{ObjectDetector}(I^{t-h:t})$, $\O$ include dynamic and static masks $D,\S$
		\STATE Instance masks  $X^{t-h:t} \gets \text{InstanceLocalization}(I^{t-h:t},D^{t-h:t})$
		\STATE Predicted instance masks $\hat{X}^{t+1} \gets \varnothing$
		\FOR{each instance mask $x$ in $X$}
		\STATE Effects from spatial relations $m_{\text{1}}^{t} \gets  \text{RelationNet}(x^{t},\O^{t},a^{t})$
		\STATE Effects from temporal relations $m_{\text{2}}^{t} \gets \text{InertiaNet}(x^{t-h:t},a^{t})$
		\STATE Total effects $m^{t} \gets m_{\text{1}}^{t}+m_{\text{2}}^{t}$
		\STATE Predicted instance mask $\hat{x}^{t+1} \gets \text{Transformation}(x^{t},m^{t})$
		\STATE $\hat{X}^{t+1} \gets \hat{X}^{t+1} \bigcup \hat{x}^{t+1}$
		\ENDFOR
		\STATE Background image $B^{t+1} \gets \text{BackgroundConstructor}(I^{t},\S^{t})$
		\STATE Predicted next frame $\hat{I}^{t+1} \gets \text{Merge}(\hat{X}^{t+1},B^{t+1})$
	\end{algorithmic}
	\label{paradigm}
\end{algorithm*}

\begin{algorithm*}[htbp]
	\caption{Region Proposal Sampling. } \label{alrps}
	\hspace*{0.02in} {\bf Input:}
	Dynamic object mask $D \in [0, 1]^{H\times W}$, the number of region proposal scales $n_S$, the folds of full coverage $T$ .
	
	\begin{algorithmic}[1]
		\STATE Initialize proposal set $\mathbb{P}=\varnothing$.
		\STATE Binarize $D$ to get the indicator for the existence of objects
		\FOR {$l=1\dots n_S$}
		\STATE Select scale $dx,dy$ depend on the level $l$.
		\FOR {$t=1\dots T$}
		\STATE Initialize candidate set $\mathbb{C} = \{(i,j) | D_{i,j}=1\}$.
		\WHILE {$\mathbb{C}\neq\varnothing$}
		\STATE Sample a pixel coordinate $(x,y)$ from $\mathbb{C}$.
		\STATE Get a box $\mathbb{B}=\{(i,j)|~|i-x|\le dx,|j-y|\le dy\}$.
		\IF {$\mathbb{B}$ is not empty}
		\STATE Insert $\mathbb{B}$ into the proposal set $\mathbb{P}\gets\mathbb{P}\cup\{\mathbb{B}\}$.
		\ENDIF
		\STATE Update the remain candidate set $\mathbb{C}\gets\mathbb{C}\setminus\mathbb{B}$.
		\ENDWHILE
		\ENDFOR
		\ENDFOR
		
	\end{algorithmic}
	
	{\bf Return:} $\mathbb{P}$
	\label{RPS}
\end{algorithm*}

\begin{table}[htp]

	\caption{Long-range prediction of MAOP in unseen environments on \emph{Monster Kong}. MAOP is trained for 1-step prediction in 3 environments and tested for 3-step and 6-step prediction in 5 unseen environments.}
	\smallskip
	\centering
	\begin{tabular}{c   c @{~} c | c @{~} c | c @{~} c }
		\toprule
		
		& \multicolumn{2}{c}{0-acc} & \multicolumn{2}{c}{1-acc} & \multicolumn{2}{c}{2-acc}\\
		\cmidrule(r){2-7}
		&  Agent & All & Agent & All & Agent & All \\
		\midrule
		3-steps &  {0.81} & {0.81} & {0.89} & {0.87} & {0.93} & {0.90} \\

		\midrule
		
		6-steps &  {0.50} & {0.53} & {0.66} & {0.67} & {0.74} & {0.73} \\
		\bottomrule
	\end{tabular}
	
	\label{long}

\end{table}

\begin{table}[htp]
	\caption{Average motion prediction error in two experiment environments. $\dag$, $\ddag$ and $\S$ indicate training with only $100$, $300$ and $3000$ samples. ALL represents all dynamic objects. }
	\smallskip
	\begin{tabular}{c   c c c c | c  c }
		\toprule
		\multirow{2}*{Model}& \multicolumn{4}{c}{Monster Kong} & \multicolumn{2}{c}{Flappy Bird\qquad\quad}\\
		\cmidrule(r){2-7}
		& 1-5$^\S$ & 1-5 & 2-5 & 3-5 & 1-5$^\dag$ & 1-5$^\ddag$\\
		\midrule
		MAOP & 0.34 & 0.15 & 0.14 & 0.12 & 0.30 & 0.34\\
		\bottomrule
	\end{tabular}
	\centering

	\label{tab:motion}
\end{table}

\begin{table}[htp]
	\caption{Accuracy of our discovery of controllable agent in unseen environments.}
	\smallskip
	\begin{tabular}{c   c c c }
		\toprule
		Model& MonsterKong & FlappyBird & Freeway \\
		\midrule
		MAOP& 100\% & 100\% & 98.75\%\\
		\bottomrule
	\end{tabular}
	\centering

	\label{control}
\end{table}

\begin{table*}[htbp]
	\caption{Prediction performance of the dynamic instance level with different region proposals of dynamic instances on 3-to-5 generalization problem of \emph{Monster Kong}. "All" represents all dynamic objects. "Computed by DIS" refers to using the proposal regions of dynamic instances computed from the level of dynamic instance segmentation in MAOP.}
	\smallskip
	\centering
	\begin{tabular}{c   c @{~} c | c @{~} c | c @{~} c | c @{~} c | c @{~} c | c @{~} c }
		\toprule
		\multirow{3}*{Noise type of proposals}& \multicolumn{6}{c}{Training environments} & \multicolumn{6}{c}{Unseen environments}\\
		\cmidrule(r){2-13}
		& \multicolumn{2}{c}{0-acc} & \multicolumn{2}{c}{1-acc} & \multicolumn{2}{c}{2-acc}& \multicolumn{2}{c}{0-acc} & \multicolumn{2}{c}{1-acc} & \multicolumn{2}{c}{2-acc}\\
		\cmidrule(r){2-13}
		& Agent & All & Agent & All & Agent & All & Agent & All & Agent & All & Agent & All \\
		\midrule
		Computed by DIS & 0.99 & 0.95 & 1.00 & 0.97 & 1.00 & 0.97 & 0.99 & 0.94&  1.00 & 0.96 &  1.00 & 0.97 \\
		Gaussian Noise & 0.63 & 0.57 & 0.94 & 0.89 & 0.99 & 0.95 & 0.60 & 0.57 & 0.93 & 0.89 & 0.98 & 0.95\\
		Poisson Noise& 0.93 & 0.91 & 0.98 & 0.95 & 0.99 & 0.96 & 0.93 & 0.91 & 0.99 & 0.96 & 0.99 & 0.96\\
		\bottomrule
	\end{tabular}
	
	\label{noiseres}

\end{table*}

\begin{table*}[htp]
	
	\caption{Performance (0-error accuracy) of MAOP with different number of object masks. }
	\label{mn}
	\smallskip
	\centering
	
		\begin{tabular}{c  cc | cc }
			\toprule
			\multirow{2}*{Mask Number}& \multicolumn{2}{c|}{Training} & \multicolumn{2}{c}{Unseen}\\
			\cmidrule(r){2-5}
			&Agent & All  & Agent & All\\
			\midrule
			3 masks  & 0.92 & 0.91  & 0.90  & 0.88  \\
			%\cline{2-7}
		    5 masks & 0.95 & 0.92 & 0.94 & 0.90  \\
			%\cline{2-7}
			8 masks & 0.93 & 0.90 & 0.91 & 0.88 \\
			\bottomrule
		\end{tabular}

\end{table*}

\end{document}